\documentclass[11pt]{article}

\usepackage[preprint]{acl}

\usepackage{times}
\usepackage{latexsym}
\usepackage{amssymb}
\usepackage{amsmath}

\usepackage[T1]{fontenc}

\usepackage[utf8]{inputenc}

\usepackage{microtype}

\usepackage{inconsolata}

\usepackage{graphicx}

\usepackage{multirow}
\usepackage{booktabs}
\usepackage{array}
\usepackage{adjustbox}

\usepackage{tabularx}
\usepackage{booktabs}
\usepackage{enumitem}

\usepackage{float}

\usepackage{amsmath,amssymb}
\usepackage{booktabs}
\usepackage{algorithm}
\usepackage{algorithmic}
\usepackage{tikz}
\usetikzlibrary{arrows.meta,positioning,shapes.geometric,calc}

\title{Enhancing Language Models for Robust Greenwashing Detection}

\author{
  \textbf{Neil Heinrich Braun\textsuperscript{1}},
  \textbf{Keane Ong\textsuperscript{1,2}},
  \textbf{Rui Mao\textsuperscript{3}},
  \textbf{Erik Cambria\textsuperscript{3}},
  \textbf{Gianmarco Mengaldo\textsuperscript{1}}
\\
\\
  \textsuperscript{1}National University of Singapore,
  \textsuperscript{2}Massachusetts Institute of Technology,
  \textsuperscript{3}Nanyang Technological University
}

\begin{document}
\maketitle
\begin{abstract}
Sustainability reports are critical for ESG assessment, yet greenwashing and vague claims often undermine their reliability. Existing NLP models lack robustness to these practices, typically relying on surface-level patterns that generalize poorly. We propose a parameter-efficient framework that structures LLM latent spaces by combining contrastive learning with an ordinal ranking objective to capture graded distinctions between concrete actions and ambiguous claims. Our approach incorporates gated feature modulation to filter disclosure noise and utilizes MetaGradNorm to stabilize multi-objective optimization. Experiments in cross-category settings demonstrate superior robustness over standard baselines while revealing a trade-off between representational rigidity and generalization.\footnotemark 
\end{abstract}

\footnotetext{Code will be released upon review.}

\section{Introduction}
Sustainability reports are key to evaluating corporate ESG performance, but their credibility is increasingly undermined by greenwashing, where vague or inflated claims blur the line between real action and rhetoric~\citep{pimonenko2020greenwashing}. Although NLP systems are widely used to analyze these disclosures, they remain vulnerable to strategically crafted language~\citep{ong2025a3cg}. Current models often over-rely on surface-level lexical and stylistic patterns, leading to strong in-distribution performance but poor generalization to unseen categories or shifted formulations. Crucially, increasing model capacity does not consistently improve robustness in this setting. This limitation is particularly problematic for ESG analysis, as companies can strategically alter their reporting focus or linguistic framing without changing underlying sustainability practices, requiring models that can look beyond surface-level cues to identify substantive actions~\citep{stammbach2023environmental}.

A key source of model fragility is that ESG analysis is typically framed as a flat prediction problem, optimizing a single objective without imposing structure on the latent representation space~\citep{ong2025a3cg}. Consequently, critical distinctions such as degrees of actionability, commitment, or evidential grounding are not explicitly encoded. Since greenwashing emerges from nuanced combinations of intent and substantiation rather than isolated cues, we hypothesize that robustness can be improved by explicitly structuring latent representations during fine-tuning~\citep{stoehr2023sentiment}. By explicitly encoding ordinal relationships between concrete actions and ambiguous claims, models may better capture underlying intent. Yet, enforcing such structure introduces a trade-off where rigid constraints potentially limit generalization to unseen cases.

We propose a parameter-efficient framework that applies structured representation learning directly to low-rank adapters (LoRA). Our approach integrates a contrastive objective to cluster semantically related claims with an ordinal ranking loss to enforce graded distinctions of actionability. A gated feature modulation mechanism further mitigates noise by dynamically emphasizing task-relevant dimensions. To stabilize the resulting multi-loss optimization, we introduce MetaGradNorm, which automatically balances competing gradients and loss contributions to ensure robust convergence without manual tuning~\citep{chen2018gradnorm}.

We evaluate our framework on the A3CG dataset~\citep{ong2025a3cg} using a cross-category generalization setting across several open-source LLMs: T5~\citep{raffel2020t5}, LLaMA-3-8B~\citep{dubey2024llama3}, Mistral-7B~\citep{jiang2023mistral}, Gemma-7B~\citep{team2024gemma}, DeepSeek-V3-7B~\citep{deepseek2024v3}, and Qwen2.5-7B~\citep{yang2024qwen2}. Our key findings include: (1) While T5 achieves the strongest overall performance, our framework consistently improves robustness across all other LLMs. (2) For these medium-scale models, structured fine-tuning yields higher robustness than 70B-parameter variants and proprietary systems, suggesting that representation structure is more critical than model scale. (3) Despite these gains, simple LoRA remains a highly competitive and consistent baseline across settings. (4) Finally, we discuss remaining limitations and outline future research strategies.

Our main contributions are as follows: (1) A structured PEFT framework that integrates contrastive and ordinal representation learning directly into low-rank adapters for robust ESG claim analysis. (2) A gated feature modulation mechanism to filter noise and strategic language in sustainability disclosures by dynamically emphasizing task-relevant dimensions. (3) MetaGradNorm, an optimization strategy that adaptively balances competing objectives to stabilize multi-loss training in structured representation settings. (4) A comprehensive empirical evaluation of robustness and generalization trade-offs across multiple LLMs and ESG categories, identifying key scaling and structural limitations.

\section{Related Work}

\paragraph{NLP for ESG Analysis and Greenwashing}
While NLP has been extensively applied to ESG disclosure analysis for tasks such as sustainability topic extraction~\citep{ong2025esgsentinet} and structured information extraction~\citep{song2018sustainable, bronzini2024glitter}, the strategic ambiguity inherent in greenwashing remains a fundamental challenge~\citep{ong2024explainable}. Recent shifts toward the Aspect–Action Modeling and Cross-Category Generalization (A3CG) framework address this by linking ESG aspects to specific corporate actions categorized along an ordinal continuum of commitment - ranging from indeterminate to implemented~\citep{ong2025a3cg}. Despite the rise of Large Language Models (LLMs) in this domain, scaling alone has proven insufficient for fine-grained sustainability analysis; proprietary models often struggle with the cross-category generalization and linguistic sensitivity required to distinguish rhetorical patterns from credible actions~\citep{ong2025a3cg, bronzini2024glitter}.
\paragraph{Parameter-Efficient Fine-Tuning for LLMs.}
While methods like LoRA~\citep{hu2021lora} reduce computational costs by adapting models via low-rank matrices~\citep{han2024peft, lialin2023scaling}, they typically rely on standard supervised objectives that do not explicitly structure the representation space. Consequently, these techniques often struggle with robustness and generalization in complex tasks due to a lack of mechanisms for encoding task-aligned constraints during adaptation.
\paragraph{Contrastive Learning for NLP Representation Learning}
Contrastive learning enhances representation quality by clustering semantically similar instances while separating dissimilar ones, improving robustness in sentence and aspect-based tasks~\citep{wang2020understanding,gao2021simcse}. However, its effectiveness depends heavily on hyperparameter tuning (e.g., margin, temperature). Furthermore, rigid contrastive constraints can hinder generalization if the training data's semantic structure fails to align with unseen distributions~\citep{huang2023model}.
\paragraph{Ordinal and Structured Objectives in NLP}
Beyond contrastive learning, ordinal regression and ranking-based losses capture graded relationships - such as sentiment intensity or stance - that discrete labels miss~\citep{stoehr2023sentiment,zhao2016ordinal}. While these objectives improve interpretability and alignment by encoding relative distinctions, integrating them into multi-objective pipelines is difficult. Furthermore, enforcing such structures can introduce rigidity, potentially limiting performance on ambiguous or unseen cases~\citep{baly2019multitask}.
\paragraph{Gating Mechanisms for Feature Modulation}
Gating mechanisms dynamically modulate information flow by selectively emphasizing task-relevant features, which improves robustness in NLP tasks with noisy or heterogeneous inputs — such as aspect-based sentiment analysis and domain adaptation~\citep{xue2018aspect,zhang2020improving}. In ESG analysis, where boilerplate obscures signals, feature-level gating offers a principled way to attenuate irrelevant dimensions while amplifying representations of meaningful sustainability actions.~\citep{feng2023ssemgat}.
\paragraph{Multi-Objective Optimization and Loss Balancing}
Multi-objective training often suffers from competing gradients and imbalanced dynamics, where fixed weighting schemes allow a single objective to dominate or destabilize training~\citep{chen2018gradnorm,yu2020gradient}. To mitigate this, adaptive methods like GradNorm dynamically adjust weights based on gradient magnitudes, while meta-learning extensions further stabilize learning~\citep{liu2022metaweighting}. These strategies are essential in frameworks combining supervised, contrastive, and structured objectives to ensure balanced convergence across different loss scales.

\section{Methodology}

\subsection{Task Definition and Problem Setup}
We study ESG claim analysis under an aspect–action formulation, where each claim is represented as a tuple pairing a sustainability aspect with an associated action description. The action component captures how the aspect is operationalized, from abstract commitments to concrete, verifiable actions. Following prior work on aspect–action modeling~\citep{ong2025a3cg}, actions are organized along an ordinal continuum of increasing actionability. Each claim is assigned an ordinal label $y \in \{\textit{indeterminate}, \textit{planning}, \textit{implemented}\}$, reflecting progressively stronger evidential grounding. The objective is to learn representations that preserve these ordinal relationships while generalizing across unseen ESG categories.

\subsection{Embedding Extraction and Similarity Functions}

Let $f_\theta$ denote a pretrained language model with parameters $\theta$. Given an input claim $x$, we extract a latent embedding $x \in \mathbb{R}^d$, which is L2-normalized as:
\begin{equation}
\hat{x} = \frac{x}{\|x\|_2}.
\end{equation}

We define cosine similarity and distance as:
\begin{equation}
\mathrm{sim}(u, v) = \hat{u}^\top \hat{v}, \quad \hat{d}(u, v) = 1 - \mathrm{sim}(u, v),
\end{equation}
which are used consistently across all objectives.

\subsection{Parameter-Efficient Adaptation with LoRA}

We employ Low-Rank Adaptation (LoRA)~\citep{hu2021lora} and use the same LoRA adapters across both structured representation learning and subsequent PEFT fine-tuning, ensuring continuity of the learned latent space. Adapters are injected into the attention projections and the feed-forward projections.

\subsection{Contrastive Representation Learning}
To structure the latent space, we employ supervised contrastive learning~\citep{khosla2020supervised}. For each anchor embedding $a$, we consider a set of positive samples $\{p_k\}_{k=1}^{K}$ and negative samples $\{n_m\}_{m=1}^{M}$. Using cosine similarity $\mathrm{sim}(\cdot,\cdot)$ and temperature $\tau > 0$, we define the multi-positive contrastive loss as:
\begin{equation}
\label{eq:contrastive}
\resizebox{0.9\columnwidth}{!}{$\displaystyle
\mathcal{L}^{(i)}_{\text{ctr}} = -\log \frac{\sum_{k=1}^{K} \exp(\mathrm{sim}(a, p_k)/\tau)} {\sum_{k=1}^{K} \exp(\mathrm{sim}(a, p_k)/\tau) + \sum_{m=1}^{M} \exp(\mathrm{sim}(a, n_m)/\tau)}
$}
\end{equation}
This objective encourages semantically related claims to cluster in the latent space while separating unrelated ones.

\subsection{Ordinal Ranking Objective}

While contrastive learning enforces similarity structure, it does not explicitly encode ordinal relationships between action levels. To address this limitation, we introduce an ordinal ranking objective aligned with the actionability continuum in A3CG~\citep{stoehr2023sentiment}.

For each anchor, we define an ordinal mean-margin loss with a constant margin $m_0 > 0$:
\begin{equation}
\label{eq:ordinal}
\resizebox{0.9\columnwidth}{!}{$\displaystyle
\mathcal{L}^{(i)}_{\text{ord}} = \max\!\left( 0, \frac{1}{K}\sum_{k=1}^{K} \hat{d}(a, p_k) - \frac{1}{M}\sum_{m=1}^{M} \hat{d}(a, n_m) + m_0 \right)
$}
\end{equation}

This loss enforces that higher-actionability claims remain closer to the anchor than lower-actionability ones by a fixed margin, aligning representation geometry with the ordinal action scale.

\subsection{Per-Sample Gating Mechanism}

To balance contrastive and ordinal objectives at the sample level, we introduce a gating mechanism with temperature parameters $T_{\text{ctr}}, T_{\text{ord}} > 0$. For each sample $i$, we compute:
\begin{equation}
\label{eq:gating_scores}
s^{(i)}_{\text{ctr}} = \frac{\mathcal{L}^{(i)}_{\text{ctr}}}{T_{\text{ctr}}} - \frac{\mathcal{L}^{(i)}_{\text{ord}}}{T_{\text{ord}}}, \quad s^{(i)}_{\text{ord}} = \frac{\mathcal{L}^{(i)}_{\text{ord}}}{T_{\text{ord}}} - \frac{\mathcal{L}^{(i)}_{\text{ctr}}}{T_{\text{ctr}}}.
\end{equation}

Normalized gating weights are obtained via:
\begin{equation}
\label{eq:gating_weights}
(w^{(i)}_{\text{ctr}}, w^{(i)}_{\text{ord}}) = \mathrm{softmax}(s^{(i)}_{\text{ctr}}, s^{(i)}_{\text{ord}}), \quad w^{(i)}_{\text{ctr}} + w^{(i)}_{\text{ord}} = 1.
\end{equation}

\subsection{Multi-Objective Learning and Loss Balancing}
The batch-level objective is defined as:
\begin{equation}
\label{eq:batch_loss}
\resizebox{0.9\columnwidth}{!}{$\displaystyle
\mathcal{L}(\theta; \alpha) = \frac{1}{B}\sum_{i=1}^{B} \left[ \lambda_{\text{base}} w^{(i)}_{\text{ctr}} \mathcal{L}^{(i)}_{\text{ctr}} + \lambda_{\text{ord}} w^{(i)}_{\text{ord}} \mathcal{L}^{(i)}_{\text{ord}} \right]
$}
\end{equation}
where $\alpha = \{\lambda_{\text{base}}, \lambda_{\text{ord}}, T_{\text{ctr}}, T_{\text{ord}}\}$ are meta-parameters.

\subsection{MetaGradNorm for Loss Balancing}
To stabilize multi-objective training, we adopt a MetaGradNorm strategy derived from GradNorm~\citep{chen2018gradnorm}. 
Gradient norms are computed as:
\begin{equation}
\label{eq:grad_norms}
\begin{split}
G_{\text{ctr}} &= \left\| \nabla_\theta \left( \frac{1}{B}\sum_i \lambda_{\text{base}} w^{(i)}_{\text{ctr}} \mathcal{L}^{(i)}_{\text{ctr}} \right) \right\|_2, \\
G_{\text{ord}} &= \left\| \nabla_\theta \left( \frac{1}{B}\sum_i \lambda_{\text{ord}} w^{(i)}_{\text{ord}} \mathcal{L}^{(i)}_{\text{ord}} \right) \right\|_2.
\end{split}
\end{equation}
Normalized losses define difficulty ratios:
\begin{equation}
\label{eq:difficulty_ratios}
\tilde{L}_k(t) = \frac{L_k(t)}{L_k(0) + \varepsilon}, \quad r_k(t) = \left( \frac{\tilde{L}_k(t)} {\frac{1}{2}\sum_j \tilde{L}_j(t)} \right)^{\gamma},
\end{equation}
where $\gamma > 0$ controls the sensitivity to task difficulty imbalance; higher values amplify gradient adjustments for harder tasks.
Target gradient norms are:
\begin{equation}
\label{eq:target_grad}
\bar{G} = \frac{1}{2}(G_{\text{ctr}} + G_{\text{ord}}), \quad G_k^\star = \bar{G}\, r_k(t).
\end{equation}
The meta-objective is:
\begin{equation}
\label{eq:meta_objective}
\resizebox{0.9\columnwidth}{!}{$\displaystyle
\mathcal{J}(\alpha \mid \theta) = |G_{\text{ctr}} - G^\star_{\text{ctr}}| + |G_{\text{ord}} - G^\star_{\text{ord}}| + \beta \mathcal{R}_{\text{ent}}(\alpha)
$}
\end{equation}
where $\beta \geq 0$ controls the strength of entropy regularization, encouraging balanced gating weights across samples. The entropy term is defined as:
\begin{equation}
\label{eq:entropy_reg}
\resizebox{0.9\columnwidth}{!}{$\displaystyle
\mathcal{R}_{\text{ent}}(\alpha) = -\frac{1}{B}\sum_{i=1}^{B} \big( w^{(i)}_{\text{ctr}}\log w^{(i)}_{\text{ctr}} + w^{(i)}_{\text{ord}}\log w^{(i)}_{\text{ord}} \big)
$}
\end{equation}

\subsection{Optimization Procedure}
Model parameters and meta-parameters are updated jointly as:
\begin{equation}
\label{eq:optimization}
\theta \leftarrow \theta - \eta_\theta \nabla_\theta \mathcal{L}(\theta; \alpha), \quad \alpha \leftarrow \alpha - \eta_\alpha \nabla_\alpha \mathcal{J}(\alpha \mid \theta),
\end{equation}
with positivity enforced via softplus reparameterization.

\subsection{Training Procedure}
Training is conducted in two stages using the same set of LoRA adapters. In the first stage, structured representation learning is performed by optimizing contrastive and ordinal objectives with gated representations and MetaGradNorm-based loss balancing, shaping the latent space to reflect graded actionability. In the second stage, task-specific fine-tuning is applied to the same adapters without reinitialization, preserving the learned structure while adapting the model to downstream supervision.

\section{Experiments}

\subsection{Experimental Setup}

We conduct experiments on the A3CG dataset~\citep{ong2025a3cg}, where each sample $x_i$ is associated with a set of tuples:
\begin{equation}
\label{eq:label_set}
\mathcal{L}_i \subseteq \mathcal{K} \times \mathcal{A},
\end{equation}
with $\mathcal{K}$ denoting sustainability aspects or categories, and $\mathcal{A} = \{\textit{indeterminate}, \textit{planning}, \textit{implemented}\}$ representing ordinal action levels. Evaluation follows a cross-category generalization protocol with disjoint training and test categories.

\paragraph{Contrastive Pair Construction.}
For each anchor $i$ with labels $\mathcal{L}_i$, we define the intersection $I_{ij} = \mathcal{L}_i \cap \mathcal{L}_j$ with candidate $j$. Positive and negative sets are:
\begin{equation}
\label{eq:pos_neg_sets}
\begin{split}
\mathcal{P}(i) &= \{\, j \neq i \mid |I_{ij}| > 0 \,\}, \\
\mathcal{N}(i) &= \{\, j \neq i \mid |I_{ij}| = 0 \,\}.
\end{split}
\end{equation}

\paragraph{Ordinal Pair Construction.}
Each $\mathcal{L}_i$ is mapped to a dictionary $d_i : k \mapsto a$. For shared keys $\mathcal{K}_{ij} = \mathrm{keys}(d_i) \cap \mathrm{keys}(d_j)$, ordinal relationships are determined via a transition function:
\begin{equation}
\label{eq:transition_function}
\pi(a)= \begin{cases} 
\texttt{planning} & \text{if } a=\texttt{implemented},\\ 
\texttt{implemented} & \text{if } a=\texttt{planning},\\ 
\texttt{planning} & \text{if } a=\texttt{indeterminate}. 
\end{cases}
\end{equation}
A candidate $j$ is a positive ordinal sample if $\exists\, k \in \mathcal{K}_{ij}$ such that $d_j(k) = \pi(d_i(k))$; all others are negatives.

\subsection{Models}

We evaluate our approach on a diverse set of pretrained open-source language models: T5~\citep{raffel2020t5}, LLaMA-3-8B~\citep{touvron2024llama3}, Mistral-7B~\citep{jiang2023mistral}, Gemma-7B~\citep{team2024gemma}, DeepSeek-7B~\citep{liu2024deepseek}, and Qwen2.5-7B~\citep{yang2024qwen2}. All models are used with frozen backbones and adapted exclusively via LoRA adapters following the same training protocol. Representations are extracted uniformly across models, and the same adapters are optimized during both structured representation learning and subsequent task-specific fine-tuning.

\subsection{Baselines and Ablation Study}

We evaluate stability and cross-category generalization using a multi-fold protocol. First, we establish a control setting using standard balanced splits on the full dataset. For generalization, we adopt the A3CG three-fold protocol~\citep{ong2025a3cg}, where specific aspect categories are withheld from training to form an unseen (US) test set, while disjoint samples from training categories form the seen (S) test set.

To isolate component contributions, we evaluate an increasingly expressive sequence of configurations. Starting from a LoRA-based PEFT baseline~\citep{hu2021lora}, we incrementally integrate contrastive learning~\citep{khosla2020supervised}, ordinal supervision, gated feature modulation, and MetaGradNorm~\citep{chen2018gradnorm}. All models are trained under identical conditions and evaluated at 4 and 6 epochs. This cross-fold strategy ensures a controlled analysis of robustness and performance gains across varying training regimes. 

\subsection{Evaluation Metrics}
We use F1-score as the primary metric, focusing on AAA, and report results separately on seen (S) and unseen (US) test sets to evaluate generalization. To ensure stability, results are averaged across the three A3CG folds.  

\subsection{Implementation Details and Hyperparameter Tuning}
We employ a staged tuning strategy on a fixed model and fold to ensure stability. We first calibrate the LoRA configuration and learning rate to establish a baseline~\citep{schulman2025lora}. We then independently tune the contrastive learning rate~\citep{gao2021simcse} and the ordinal margin $m_0$~\citep{su2017deep} to balance structure and generalization. Subsequently, gating temperatures $T_{\text{ctr}}, T_{\text{ord}}$ and scaling parameters $\lambda_{\text{ctr}}, \lambda_{\text{ord}}$ are optimized to regulate relative loss influence. Finally, we tune MetaGradNorm parameters $\gamma$ and $\beta$ to stabilize multi-objective convergence~\citep{chen2018gradnorm}. Final hyperparameters are fixed and applied consistently across all models and folds to ensure reproducibility and fair comparison.  

\section{Results and Discussion}

We evaluate our framework on the AAA task, prioritizing robustness and cross-category generalization over in-distribution performance. Results are reported via F1-score on seen (S) and unseen (US) categories (Section 4).

\begin{table*}[ht]
\centering
\caption{T5 Results: Comparison between our method and A3CG baseline}
\adjustbox{max width=\textwidth}{
\begin{tabular}{|l|c|cc|cc|cc|c|c|c|}
\hline
\multirow{2}{*}{\textbf{Method}} & \textbf{Full} & \multicolumn{2}{c|}{\textbf{Fold 1}} & \multicolumn{2}{c|}{\textbf{Fold 2}} & \multicolumn{2}{c|}{\textbf{Fold 3}} & \multirow{2}{*}{\textbf{S Avg}} & \multirow{2}{*}{\textbf{US Avg}} & \multirow{2}{*}{$\Delta$} \\
\cline{3-8}
 & \textbf{Dataset} & S & US & S & US & S & US & & & \\
\hline
T5  & 71.60 & 56.76 & 48.63 & 66.83 & \textbf{\underline{49.47}} & \textbf{\underline{69.94}} & 35.89 & 64.51 & 44.66 & -19.85 \\
T5 + COGLM  & \textbf{\underline{72.40}} & \textbf{\underline{63.64}} & \textbf{\underline{52.11}} & \textbf{\underline{69.97}} & 48.57 & 69.15 & \textbf{\underline{46.44}} & \textbf{\underline{67.59}} & \textbf{\underline{49.04}} & -18.55 \\
\hline
T5 (A3CG) & 70.48 & 57.85 & 43.03 & 68.90 & 45.74 & 67.94 & 34.59 & 64.90 & 41.12 & -23.78 \\
T5 + CL (A3CG) & \underline{71.12} & \underline{62.96} & \underline{46.97} & \underline{69.76} & \underline{46.67} & \underline{67.99} & \underline{38.33} & \underline{66.90} & \underline{43.99} & -22.91 \\
\hline
\end{tabular}
}
\label{tab:t5_results}
\end{table*}

\begin{table*}[ht]
\centering
\caption{LLM Results: Fine-tuned models (LoRA)}
\adjustbox{max width=\textwidth}{
\begin{tabular}{|l|c|cc|cc|cc|c|c|c|}
\hline
\multirow{2}{*}{\textbf{Method}} & \textbf{Full} & \multicolumn{2}{c|}{\textbf{Fold 1}} & \multicolumn{2}{c|}{\textbf{Fold 2}} & \multicolumn{2}{c|}{\textbf{Fold 3}} & \multirow{2}{*}{\textbf{S Avg}} & \multirow{2}{*}{\textbf{US Avg}} & \multirow{2}{*}{$\Delta$} \\
\cline{3-8}
 & \textbf{Dataset} & S & US & S & US & S & US & & & \\
\hline
\multicolumn{11}{|c|}{\textit{Fine-tuned (LoRA)}} \\
\hline
LLaMA-3-8B & 0.5668 & 0.5095 & \underline{0.4626} & 0.6181 & 0.4422 & 0.5972 & 0.3481 & 0.5749 & 0.4176 & -0.1573 \\
LLaMA-3-8B + COGLM & 0.6075 & 0.5678 & 0.4109 & 0.6346 & 0.4566 & \textbf{\underline{0.6669}} & 0.3665 & 0.6231 & 0.4113 & -0.2118 \\
\hline
Mistral-7B & 0.5963 & \textbf{\underline{0.6191}} & 0.3880 & 0.6248 & \textbf{\underline{0.5000}} & 0.6264 & 0.3977 & \textbf{\underline{0.6234}} & \textbf{\underline{0.4286}} & -0.1948 \\
Mistral-7B + COGLM & 0.6206 & 0.5794 & 0.3786 & 0.6317 & 0.4361 & 0.6510 & 0.3864 & 0.6207 & 0.4004 & -0.2203 \\
\hline
Gemma-7B & 0.5937 & 0.5747 & 0.4510 & 0.5737 & 0.3985 & 0.6043 & 0.3580 & 0.5842 & 0.4025 & -0.1817 \\
Gemma-7B + COGLM & \textbf{\underline{0.6379}} & 0.5626 & 0.3974 & \textbf{\underline{0.6363}} & 0.4400 & 0.6170 & 0.3497 & 0.6053 & 0.3957 & -0.2096 \\
\hline
DeepSeek-7B & 0.3690 & 0.3261 & 0.3082 & 0.4952 & 0.3417 & 0.3902 & 0.3065 & 0.4038 & 0.3188 & -0.0850 \\
DeepSeek-7B + COGLM & 0.4405 & 0.4425 & 0.2953 & 0.5204 & 0.3869 & 0.4547 & 0.3053 & 0.4725 & 0.3292 & -0.1433 \\
\hline
Qwen2.5-7B & 0.5906 & 0.5203 & 0.4079 & 0.6161 & 0.4636 & 0.6109 & \textbf{\underline{0.4022}} & 0.5824 & 0.4246 & -0.1578 \\
Qwen2.5-7B + COGLM & 0.6006 & 0.5465 & 0.3440 & 0.6185 & 0.4387 & 0.6173 & 0.3902 & 0.5941 & 0.3910 & -0.2031 \\
\hline
\end{tabular}
}
\label{tab:llm_results}
\end{table*}

\subsection{Baseline vs Full Framework: Overall Robustness Trends}

Table~\ref{tab:t5_results} demonstrates the effectiveness of COGLM, which integrates contrastive learning, ordinal loss, gating, scaling with lambdas and MetaGradNorm on the T5 architecture. COGLM consistently outperforms the standard T5 and previous A3CG baselines across all folds, achieving a peak overall AAA F1 score of 0.724~\citep{ong2025a3cg}. Notably, \textbf{the framework improves unseen category performance by 4 to 5 F1 points, confirming enhanced cross-category generalization}. These results indicate that jointly structuring representations and dynamically balancing objectives provides a superior inductive bias compared to isolated components, aligning with established research in structured representation learning~\citep{kook2022deep, cheryfs2023contrastive}.

Table~\ref{tab:llm_results} demonstrates that COGLM's benefits extend to 7–8B decoder-only LLMs, consistently outperforming LoRA-only baselines. Notably, \textbf{COGLM-enhanced 7B models match or exceed the unseen-category performance of GPT-4o, Claude 3.5 Sonnet, and LLaMA-3-70B, despite having far fewer parameters}. While a gap between seen and unseen categories persists due to the complexity of ESG generalization, COGLM effectively improves performance on seen categories without any degradation in unseen results. These results confirm that structured, parameter-efficient adaptation provides a more effective inductive bias for transferable action-level representations than merely increasing model scale~\citep{ong2025a3cg, cheryfs2023contrastive}.

\begin{figure}[t]
\centering
\includegraphics[width=\columnwidth]{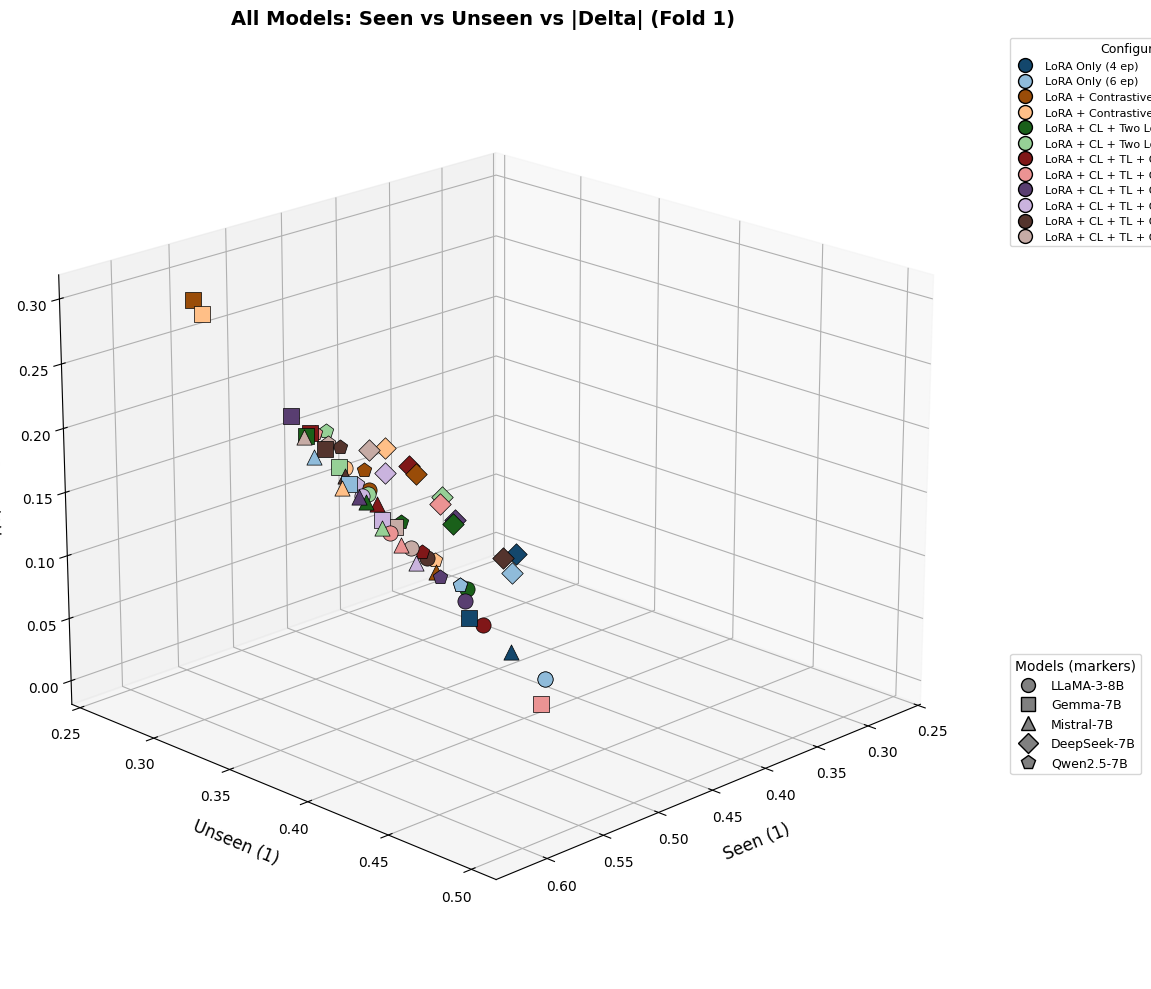}
\caption{Seen vs Unseen vs $|\Delta|$ for Fold 1. Markers denote models; colors denote configurations.}
\label{fig:trajectory}
\end{figure}

These trends are further illustrated in Figure~\ref{fig:trajectory}, which visualizes model trajectories in the joint space of seen performance, unseen performance, and $|\Delta|$ for Fold 1. As components are progressively added, models tend to move toward regions of higher unseen performance, but not always monotonically. \textbf{Robustness emerges gradually through representation structuring} rather than as a direct consequence of maximizing seen accuracy, and intermediate configurations often occupy favorable trade-off regions~\citep{ni2024peft}.

Distinct patterns emerge across model families: \textbf{encoder–decoder architectures, such as T5, tend to maintain a more stable balance between seen and unseen performance with smaller generalization gaps} dispersed trajectories, where gains in unseen performance can coincide with increased rigidity or higher variance. These findings reinforce the utility of analyzing robustness through the joint behavior of S, US, and $\Delta$ rather than relying on a single aggregate metric.

\subsection{Progressive Component Contributions and Best Configurations}

\begin{figure*}[ht]
    \centering
    \includegraphics[width=\textwidth]{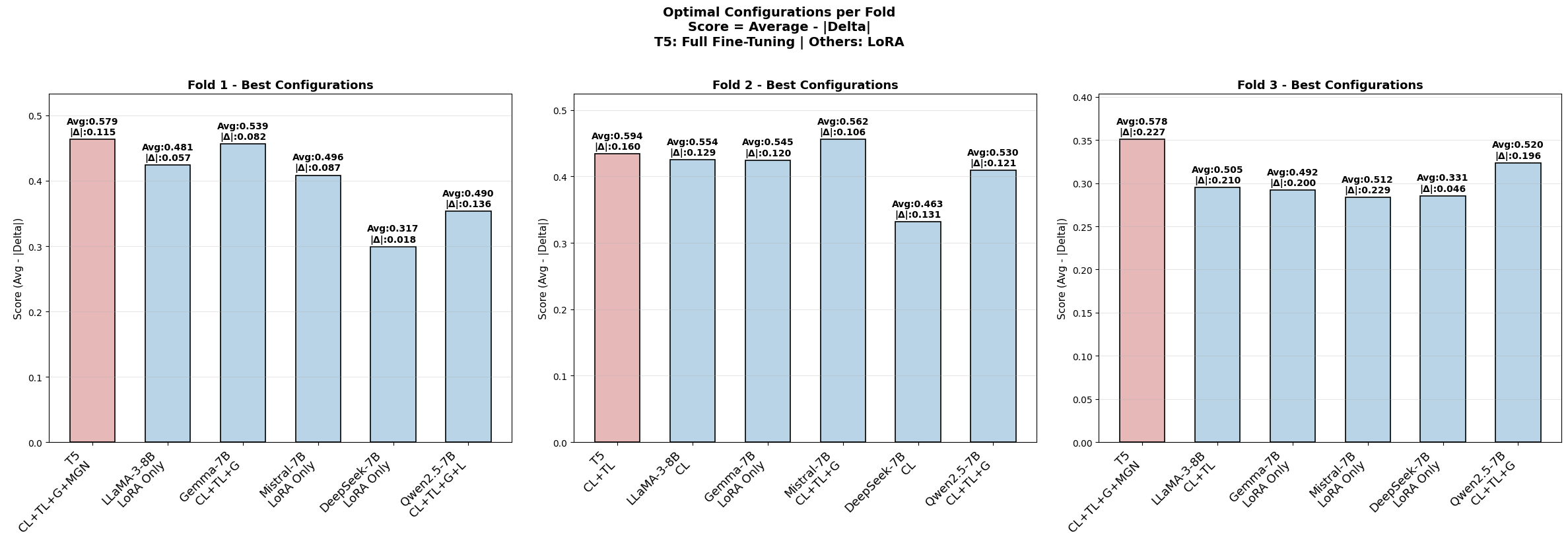}
    \caption{Best configurations comparison across folds for seen (solid) and unseen (hatched) categories. Each fold shows the F1 scores for the best-performing model on seen categories paired with its unseen performance, highlighting the generalization gap across different architectures.}
    \label{fig:seen_unseen}
\end{figure*}

Figure~\ref{fig:seen_unseen} indicates that \textbf{the most effective configuration is not necessarily the most complex}. By reporting the configuration that maximizes a composite score $\text{Avg} - |\Delta|$, we observe that robustness appears to emerge from selective component composition rather than uniform application of all objectives~\citep{zhou2024autopeft,ploner2024selective}.

Trends vary significantly across folds: in Fold 1, models like Gemma-7B and Qwen2.5-7B reach their optimal trade-off using intermediate configurations (contrastive and ordinal supervision) without additional stabilization. In contrast, the full framework is more competitive in Fold 2 for several models. Fold 3 shows the highest variability, where some models benefit from minimal structure while others require stronger constraints to manage the generalization gap~\citep{vuckovic2023kfold,nomad2023lora}. These results suggest that \textbf{the optimal structural balance is highly sensitive to the specific aspect categories encountered during training}.

Examination of component-level contributions clarifies these performance variations. While contrastive learning improves representation separation and unseen F1, it increases sensitivity to temperature and learning rates~\citep{takamoto2025temperature}. Similarly, the ordinal loss provides task-aligned structure, but its effectiveness is highly sensitive to margin choice; excessive constraints can reduce model flexibility~\citep{becerra2024ordinal}. Gated feature modulation improves stability by dynamically weighting dimensions, yet miscalibration of its scaling parameters can offset these gains~\citep{qiu2025gated}. Furthermore, MetaGradNorm balances competing gradients between contrastive and ordinal losses, though its impact varies across folds and can occasionally be less consistent than fixed-scaling mechanisms in lower-complexity settings~\citep{chen2018gradnorm}.

Crucially, each component expands the hyperparameter surface. We observe that suboptimal tuning can cause simpler pipelines to outperform more complex configurations. These findings suggest that \textbf{robustness is not a direct function of methodological complexity but rather of the calibration between structured objectives and their hyperparameters}. Overall, effective generalization appears to arise from the balanced combination of components rather than maximal complexity alone.

\subsection{Latent Space Structuring: Representation-Level Analysis}

\begin{figure*}[t]
\centering
\begin{minipage}{0.49\textwidth}
    \centering
    \includegraphics[width=\textwidth]{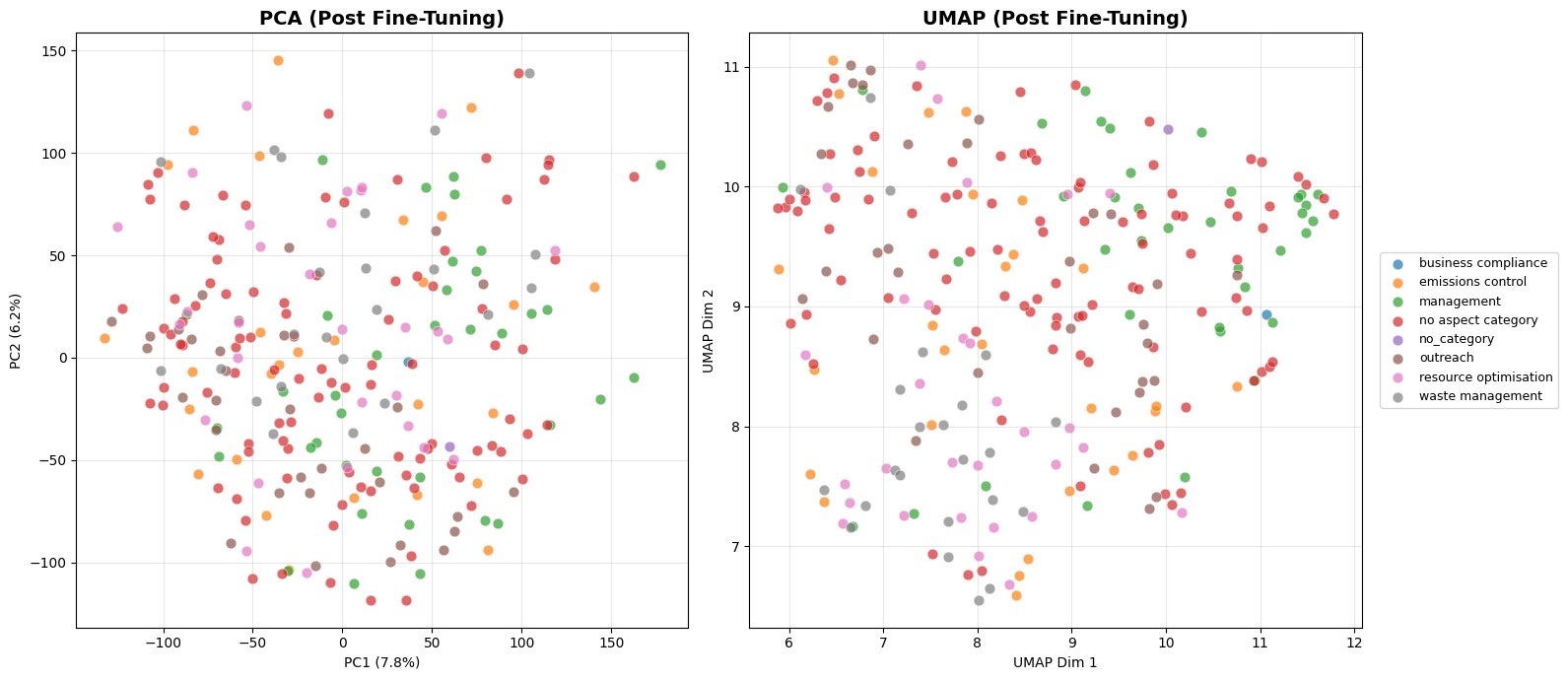}
    {\scriptsize LoRA-only baseline}\\
    {\tiny Silhouette: $-0.012$ | Calinski-Harabasz: $3.04$ | Separation: $0.47$}
    \label{fig:embedding_baseline}
\end{minipage}
\hfill
\begin{minipage}{0.49\textwidth}
    \centering
    \includegraphics[width=\textwidth]{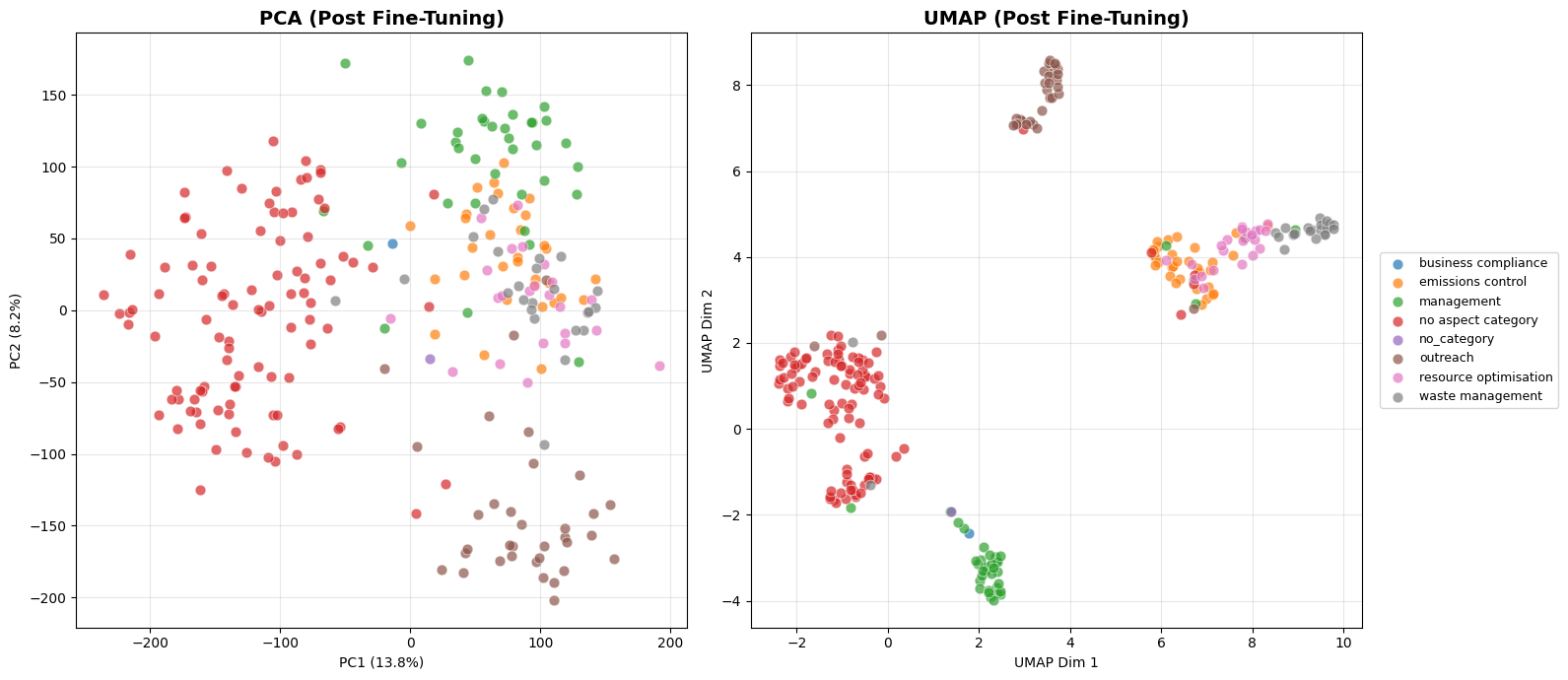}
    {\scriptsize Full structured framework}\\
    {\tiny Silhouette: $0.105$ | Calinski-Harabasz: $16.53$ | Separation: $0.73$}
    \label{fig:embedding_full}
\end{minipage}
\caption{PCA and UMAP projections of latent representations for Mistral-7B on Fold 2. Left: Baseline shows diffuse clusters with low separation. Right: Full framework yields tighter, better-separated clusters by aspect category.}
\label{fig:embeddings}
\end{figure*}

To complement our quantitative results, we examine how structured objectives influence latent space geometry. This analysis suggests that \textbf{gains in cross-category generalization are linked to shifts in embedding organization rather than optimization effects alone}~\citep{kumar2025llm}. We focus on a representative case study from Fold 2 using the most stable model identified in Section 5.2. This fold, which features diverse unseen categories, serves as an illustrative example of structured representation learning rather than a claim of uniform behavior across all settings.

Figure~\ref{fig:embeddings} compares PCA and UMAP projections of embeddings from the LoRA-only baseline and the optimal structured configuration~\citep{kobak2023umap,senties2021umap}. \textbf{In the baseline, embeddings across different aspect categories are heavily intermixed}. Specifically, claims lacking explicit aspect annotations are scattered throughout the space, suggesting limited alignment between latent representations and the underlying action-level semantics.

In contrast, \textbf{the structured configuration exhibits a more organized geometry}. Operational categories, such as emissions control and waste management, form compact regions with clearer boundaries, while managerial aspects occupy more distinct areas~\citep{vujanic2025umap}. Claims without aspect labels concentrate into a localized region rather than diffusing across the space, suggesting improved separation between actionable and non-actionable content. These patterns are particularly coherent in the UMAP projection.

This visual reorganization is supported by clustering statistics: the structured configuration yields higher silhouette and Calinski–Harabasz scores, indicating tighter intra-cluster cohesion and improved inter-cluster separation~\citep{zhang2024cl,gronbech2021scrna}. While these metrics are not directly optimized, their improvement aligns with the gains in unseen F1 on Fold 2. These findings suggest that \textbf{representation-level structuring may contribute to more transferable decision boundaries by reshaping the latent space in a manner consistent with action-level and category-level semantics}.

\subsection{Hyperparameter Sensitivity and Training Stability}

The proposed framework introduces multiple interacting objectives and control mechanisms, making \textbf{performance highly sensitive to hyperparameter choices}~\citep{zela2023nomad}. To ensure that observed gains are not incidental, we conduct a staged hyperparameter analysis, illustrated through an extensive ablation study on LLaMA-3-8B, which we use as a representative case. This analysis highlights that \textbf{stable and robust performance emerges only when hyperparameters are tuned progressively}, following the order in which components are introduced~\citep{xu2025lorapro}.

At the base level, we observe substantial variability with respect to LoRA rank, scaling factor $\alpha$, and fine-tuning learning rate. Even without additional objectives, inappropriate combinations can lead to unstable convergence or degraded unseen performance, indicating that \textbf{PEFT configuration alone already constrains the attainable robustness regime}~\citep{li2025trainsmall}.

Introducing contrastive learning further increases sensitivity. \textbf{The contrastive learning rate plays a critical role}: overly large values can cause representation collapse, while overly small values limit separation benefits~\citep{jeong2024angle}. These effects are reflected in large fluctuations in unseen F1 across otherwise similar configurations. The addition of the ordinal objective introduces another critical parameter—the margin—which directly controls the rigidity of the induced ordering. \textbf{Margins that are too large enforce overly strict separation}, leading to reduced flexibility and a systematic drop in unseen performance.

Gated feature modulation regulates latent dimension influence but introduces temperature and scaling parameters. If miscalibrated, gating may suppress useful gradients or amplify noise, potentially offsetting its intended stabilizing effect~\citep{gatera2024token}. At this stage, interactions between contrastive and ordinal objectives become fold-dependent, reflecting shifts in the semantic composition of unseen categories~\citep{m2o2025multi}.

MetaGradNorm addresses these competing pressures at the optimization level by balancing conflicting gradients, thereby reducing inter-epoch instability and variance~\citep{chen2018gradnorm,gradmultinorm2025alternating}. However, our ablation results suggest that simpler stabilization mechanisms can yield more reliable performance if MetaGradNorm is poorly configured. These findings indicate that robust structured fine-tuning requires careful, staged hyperparameter control. Ultimately, \textbf{gains emerge not from methodological complexity alone, but from the precise joint calibration of structural objectives and their associated optimization landscapes}. Appendix~D provides complete results.

\section{Conclusion}
We presented a structured, parameter-efficient framework for robust ESG claim analysis using action-level representations. On A3CG, combining contrastive and ordinal objectives with adaptive stabilization outperforms standard fine-tuning. Future work will address hyperparameter stability and transferability to reduce sensitivity across models and data splits.

\section*{Limitations}

Our study is constrained by the A3CG benchmark, which is English-only and limited to a specific taxonomy of ESG aspects and three ordinal action levels. While our framework improves robustness, we observe that gains are not monotonic across folds and models: additional components increase the hyperparameter surface and can reduce generalization when miscalibrated. Moreover, our representation analysis relies on low-dimensional projections and clustering proxies that do not fully characterize decision boundary transfer. Finally, we focus on PEFT and a fixed training protocol; alternative objectives, sampling strategies, or data augmentation could change the observed trade-offs.

\section*{Ethical Considerations}
We use publicly available corporate sustainability disclosures and follow the A3CG data governance and annotation guidelines. Although the texts may contain company or personal names, our analysis focuses on action-level content rather than individual profiling, and we avoid releasing any additional sensitive information beyond what is already public. All language models used in this study are open-source and openly accessible, selected to ensure transparency, reproducibility, and equitable access to research resources, and to avoid reliance on proprietary systems with opaque training data or constraints. We adhere to the licenses of all models, datasets, and software packages. Finally, as automated assessments of “actionability” may be misused as definitive credibility judgments, we position our framework as decision support and emphasize the need for human oversight and contextual interpretation.

\bibliography{custom}

@article{pimonenko2020greenwashing,
    author  = {Tetyana Pimonenko and Yuriy Bilan and Jakub Hor{\'a}k and Liudmyla Starchenko and Waldemar Gajda},
    title   = {Green Brand of Companies and Greenwashing Under Sustainable Development Goals},
    journal = {Sustainability},
    volume  = {12},
    number  = {4},
    pages   = {1679},
    year    = {2020}
}

@inproceedings{ong2025a3cg,
    author    = {Keane Ong and Rui Mao and Deeksha Varshney and Erik Cambria and Gianmarco Mengaldo},
    title     = {Towards Robust {ESG} Analysis Against Greenwashing Risks: Aspect-Action Analysis with Cross-Category Generalization},
    booktitle = {Proceedings of the 63rd Annual Meeting of the Association for Computational Linguistics (ACL)},
    year      = {2025}
}

@inproceedings{stammbach2023environmental,
    author    = {Dominik Stammbach and Nicolas Webersinke and Julia Anna Bingler and Mathias Kraus and Markus Leippold},
    title     = {Environmental Claim Detection},
    booktitle = {Proceedings of the 61st Annual Meeting of the Association for Computational Linguistics (Volume 2: Short Papers)},
    year      = {2023}
}

@inproceedings{stoehr2023sentiment,
    author    = {Niklas Stoehr and Ryan Cotterell and Aaron Schein},
    title     = {Sentiment as an Ordinal Latent Variable},
    booktitle = {Proceedings of the 17th Conference of the European Chapter of the Association for Computational Linguistics (EACL)},
    year      = {2023}
}

@article{raffel2020t5,
    author  = {Colin Raffel and Noam Shazeer and Adam Roberts and Katherine Lee and Sharan Narang and Michael Matena and Yanqi Zhou and Wei Li and Peter J. Liu},
    title   = {Exploring the Limits of Transfer Learning with a Unified Text-to-Text Transformer},
    journal = {Journal of Machine Learning Research},
    volume  = {21},
    number  = {140},
    pages   = {1--67},
    year    = {2020}
}

@article{dubey2024llama3,
    author  = {{Llama Team, AI @ Meta}},
    title   = {The {Llama} 3 Herd of Models},
    journal = {arXiv preprint arXiv:2407.21783},
    year    = {2024},
    note    = {Over 500 contributors}
}

@article{jiang2023mistral,
    author  = {Albert Q. Jiang and Alexandre Sablayrolles and Arthur Mensch and Chris Bamford and Devendra Singh Chaplot and Diego de las Casas and Florian Bressand and Gianna Lengyel and Guillaume Lample and Lucile Saulnier and L{\'e}lio Renard Lavaud and Marie-Anne Lachaux and Pierre Stock and Teven Le Scao and Thibaut Lavril and Thomas Wang and Timoth{\'e}e Lacroix and William El Sayed},
    title   = {Mistral 7B},
    journal = {arXiv preprint arXiv:2310.06825},
    year    = {2023}
}

@article{team2024gemma,
    author  = {{Gemma Team} and Thomas Mesnard and Cassidy Hardin and Robert Dadashi and Surya Bhupatiraju and Shreya Pathak and Laurent Sifre and Morgane Rivi{\`e}re and Mihir Sanjay Kale and Juliette Love and Pouya Tafti and L{\'e}onard Hussenot and Pier Giuseppe Sessa and Aakanksha Chowdhery and Adam Roberts and Aditya Barua and Alex Botev and Alex Castro-Ros and Ambrose Slone and Am{\'e}lie H{\'e}liou and Andrea Tacchetti and Anna Bulanova and Antonia Paterson and Beth Tsai and Bobak Shahriari and Charline Le Lan and Christopher A. Choquette-Choo and Cl{\'e}ment Crepy and Daniel Cer and Daphne Ippolito and David Reid and Elena Buchatskaya and Eric Ni and Eric Noland and Geng Yan and George Tucker and George-Christian Muraru and Grigory Rozhdestvenskiy and Henryk Michalewski and Ian Tenney and Ivan Grishchenko and Jacob Austin and James Keeling and Jane Labanowski and Jean-Baptiste Lespiau and Jeff Stanway and Jenny Brennan and Jeremy Chen and Johan Ferret and Justin Chiu and Justin Mao-Jones and Katherine Lee and Kathy Yu and Katie Millican and Lars Lowe Sjoesund and Lisa Lee and Lucas Dixon and Machel Reid and Maciej Miko{\l}ajczyk and Mateo Wirth and Michael Sharman and Nikolai Chinaev and Nithum Thain and Olivier Bachem and Oscar Chang and Oscar Wahltinez and Paige Bailey and Paul Michel and Petko Yotov and Rahma Chaabouni and Ramona Comanescu and Reena Jana and Rohan Anil and Ross McIlroy and Ruibo Liu and Ryan Mullins and Samuel L Smith and Sebastian Borgeaud and Sertan Girgin and Sholto Douglas and Shree Pandya and Siamak Shakeri and Soham De and Ted Klimenko and Tom Hennigan and Vlad Feinberg and Wojciech Stokowiec and Yu-hui Chen and Zafarali Ahmed and Zhitao Gong and Tris Warkentin and Ludovic Peran and Minh Giang and Cl{\'e}ment Farabet and Oriol Vinyals and Jeff Dean and Koray Kavukcuoglu and Demis Hassabis and Zoubin Ghahramani and Douglas Eck and Joelle Barral and Fernando Pereira and Eli Collins and Armand Joulin and Noah Fiedel and Evan Senter and Alek Andreev and Kathleen Kenealy},
    title   = {Gemma: Open Models Based on Gemini Research and Technology},
    journal = {arXiv preprint arXiv:2403.08295},
    year    = {2024}
}

@article{deepseek2024v3,
    author  = {DeepSeek-AI and Aixin Liu and Bei Feng and Bing Xue and Bingxuan Wang and Bochao Wu and Chengda Lu and Chenggang Zhao and Chengqi Deng and Chenyu Zhang and Chong Ruan and Damai Dai and Daya Guo and Dejian Yang and Deli Chen and Dongjie Ji and Erhang Li and Fangyun Lin and Fucong Dai and Fuli Luo and Guangbo Hao and Guanting Chen and Guowei Li and H. Zhang and Han Bao and Hanwei Xu and Haocheng Wang and Haowei Zhang and Honghui Ding and Huajian Xin and Huazuo Gao and Hui Li and Hui Qu and J. L. Cai and Jian Liang and Jianzhong Guo and Jiaqi Ni and Jiashi Li and Jiawei Wang and Jin Chen and Jingchang Chen and Jingyang Yuan and Junjie Qiu and Junlong Li and Junxiao Song and Kai Dong and Kai Hu and Kaige Gao and Kang Guan and Kexin Huang and Kuai Yu and Lean Wang and Lecong Zhang and Lei Xu and Leyi Xia and Liang Zhao and Litong Wang and Liyue Zhang and Meng Li and Miaojun Wang and Mingchuan Zhang and Minghua Zhang and Minghui Tang and Mingming Li and Ning Tian and Panpan Huang and Peiyi Wang and Peng Zhang and Qiancheng Wang and Qihao Zhu and Qinyu Chen and Qiushi Du and R. J. Chen and R. L. Jin and Ruiqi Ge and Ruisong Zhang and Ruizhe Pan and Runji Wang and Runxin Xu and Ruoyu Zhang and Ruyi Chen and S. S. Li and Shanghao Lu and Shangyan Zhou and Shanhuang Chen and Shaoqing Wu and Shengfeng Ye and Shengfeng Ye and Shirong Ma and Shiyu Wang and Shuang Zhou and Shuiping Yu and Shunfeng Zhou and Shuting Pan and T. Wang and Tao Yun and Tian Pei and Tianyu Sun and W. L. Xiao and Wangding Zeng and Wanjia Zhao and Wei An and Wen Liu and Wenfeng Liang and Wenjun Gao and Wenqin Yu and Wentao Zhang and X. Q. Li and Xiangyue Jin and Xianzu Wang and Xiao Bi and Xiaodong Liu and Xiaohan Wang and Xiaojin Shen and Xiaokang Chen and Xiaokang Zhang and Xiaosha Chen and Xiaotao Nie and Xiaowen Sun and Xiaoxiang Wang and Xin Cheng and Xin Liu and Xin Xie and Xingchao Liu and Xingkai Yu and Xinnan Song and Xinxia Shan and Xinyi Zhou and Xinyu Yang and Xinyuan Li and Xuecheng Su and Xuheng Lin and Y. K. Li and Y. Q. Wang and Y. X. Wei and Y. X. Zhu and Yang Zhang and Yanhong Xu and Yanhong Xu and Yanping Huang and Yao Li and Yao Zhao and Yaofeng Sun and Yaohui Li and Yaohui Wang and Yi Yu and Yi Zheng and Yichao Zhang and Yifan Shi and Yiliang Xiong and Ying He and Ying Tang and Yishi Piao and Yisong Wang and Yixuan Tan and Yiyang Ma and Yiyuan Liu and Yongqiang Guo and Yu Wu and Yuan Ou and Yuchen Zhu and Yuduan Wang and Yue Gong and Yuheng Zou and Yujia He and Yukun Zha and Yunfan Xiong and Yunxian Ma and Yuting Yan and Yuxiang Luo and Yuxiang You and Yuxuan Liu and Yuyang Zhou and Z. F. Wu and Z. Z. Ren and Zehui Ren and Zhangli Sha and Zhe Fu and Zhean Xu and Zhen Huang and Zhen Zhang and Zhenda Xie and Zhengyan Zhang and Zhewen Hao and Zhibin Gou and Zhicheng Ma and Zhigang Yan and Zhihong Shao and Zhipeng Xu and Zhiyu Wu and Zhongyu Zhang and Zhuoshu Li and Zihui Gu and Zijia Zhu and Zijun Liu and Zilin Li and Ziwei Xie and Ziyang Song and Ziyi Gao and Zizheng Pan},
    title   = {DeepSeek-V3 Technical Report},
    journal = {arXiv preprint arXiv:2412.19437},
    year    = {2024},
}

@inproceedings{chen2018gradnorm,
    author    = {Zhao Chen and Vijay Badrinarayanan and Chen-Yu Lee and Andrew Rabinovich},
    title     = {{GradNorm}: Gradient Normalization for Adaptive Loss Balancing in Deep Multitask Networks},
    booktitle = {Proceedings of the 35th International Conference on Machine Learning (ICML)},
    series    = {PMLR},
    volume    = {80},
    pages     = {1200--1209},
    year      = {2018}
}

@article{ong2025esgsentinet,
    author  = {Keane Ong and Rui Mao and Deeksha Varshney and Frank Xing and Ranjan Satapathy and Johan Sulaeman and Erik Cambria and Gianmarco Mengaldo},
    title   = {{ESGSenticNet}: A Neurosymbolic Knowledge Base for Corporate Sustainability Analysis},
    journal = {arXiv preprint arXiv:2501.15720},
    year    = {2025}
}

@article{song2018sustainable,
    author  = {Yuan Song and Hongwei Wang and Maoran Zhu},
    title   = {Sustainable Strategy for Corporate Governance Based on the Sentiment Analysis of Financial Reports with {CSR}},
    journal = {Financial Innovation},
    volume  = {4},
    pages   = {1--14},
    year    = {2018}
}

@article{ong2024explainable,
    author  = {Keane Ong and Rui Mao and Ranjan Satapathy and Ricardo Shirota Filho and Erik Cambria and Johan Sulaeman and Gianmarco Mengaldo},
    title   = {Explainable Natural Language Processing for Corporate Sustainability Analysis},
    journal = {Information Fusion},
    pages   = {102726},
    year    = {2025}
}

@article{bronzini2024glitter,
    author  = {Marco Bronzini and Carlo Nicolini and Bruno Lepri and Andrea Passerini and Jacopo Staiano},
    title   = {Glitter or Gold? Deriving Structured Insights from Sustainability Reports via Large Language Models},
    journal = {EPJ Data Science},
    volume  = {13},
    number  = {41},
    year    = {2024}
}

@article{han2024peft,
    author  = {Zeyu Han and Chao Gao and Jinyang Liu and Jeff Zhang and Sai Qian Zhang},
    title   = {Parameter-Efficient Fine-Tuning for Large Models: A Comprehensive Survey},
    journal = {Published in Transactions on Machine Learning Research },
    month   = {10},
    year    = {2024}
}

@inproceedings{hu2021lora,
    author={Edward J Hu and yelong shen and Phillip Wallis and Zeyuan Allen-Zhu and Yuanzhi Li and Shean Wang and Lu Wang and Weizhu Chen},
    title={Lo{RA}: Low-Rank Adaptation of Large Language Models},
    booktitle={International Conference on Learning Representations},
    year={2022},
    url={https://openreview.net/forum?id=nZeVKeeFYf9}
}

@article{lialin2023scaling,
    author  = {Vladislav Lialin and Vijeta Deshpande and Xiaowei Yao and Anna Rumshisky},
    title   = {Scaling Down to Scale Up: A Guide to Parameter-Efficient Fine-Tuning},
    journal = {arXiv preprint arXiv:2303.15647},
    year    = {2023}
}

@inproceedings{khosla2020supervised,
    author    = {Prannay Khosla and Piotr Teterwak and Chen Wang and Aaron Sarna and Yonglong Tian and Phillip Isola and Aaron Maschinot and Ce Liu and Dilip Krishnan},
    title     = {Supervised Contrastive Learning},
    booktitle = {Advances in Neural Information Processing Systems (NeurIPS)},
    year      = {2020}
}

@inproceedings{wang2020understanding,
    author    = {Tongzhou Wang and Phillip Isola},
    title     = {Understanding Contrastive Representation Learning through Alignment and Uniformity on the Hypersphere},
    booktitle = {Proceedings of the 37th International Conference on Machine Learning (ICML)},
    year      = {2020}
}

@inproceedings{gao2021simcse,
    author    = {Tianyu Gao and Xingcheng Yao and Danqi Chen},
    title     = {{SimCSE}: Simple Contrastive Learning of Sentence Embeddings},
    booktitle = {Proceedings of the 2021 Conference on Empirical Methods in Natural Language Processing (EMNLP)},
    year      = {2021}
}

@inproceedings{huang2023model,
    author    = {Zizheng Huang and Haoxing Chen and Ziqi Wen and Chao Zhang and Huaxiong Li and Bo Wang and Chunlin Chen},
    title     = {Model-Aware Contrastive Learning: Towards Escaping the Dilemmas},
    booktitle = {Proceedings of the 40th International Conference on Machine Learning (ICML)},
    year      = {2023}
}

@inproceedings{zhao2016ordinal,
    author    = {Yunchao He and Liang-Chih Yu and Chin-Sheng Yang and K. Robert Lai and Weiyi Liu},
    title     = {{YZU-NLP} Team at {SemEval}-2016 Task 4: Ordinal Sentiment Classification Using a Recurrent Convolutional Network},
    booktitle = {Proceedings of the 10th International Workshop on Semantic Evaluation (SemEval-2016)},
    year      = {2016}
}

@inproceedings{baly2019multitask,
    author    = {Ramy Baly and Georgi Karadzhov and Abdelrhman Saleh and James Glass and Preslav Nakov},
    title     = {Multi-Task Ordinal Regression for Jointly Predicting the Trustworthiness and the Leading Political Ideology of News Media},
    booktitle = {Proceedings of the 2019 Conference of the North American Chapter of the Association for Computational Linguistics: Human Language Technologies, Volume 1},
    pages     = {2109--2116},
    year      = {2019},
}

@inproceedings{xue2018aspect,
    author    = {Wei Xue and Tao Li},
    title     = {Aspect Based Sentiment Analysis with Gated Convolutional Networks},
    booktitle = {Proceedings of the 56th Annual Meeting of the Association for Computational Linguistics (ACL)},
    year      = {2018}
}

@inproceedings{zhang2020improving,
    author    = {Amir Pouran Ben Veyseh and Nasim Nour and Franck Dernoncourt and Quan Hung Tran and Dejing Dou and Thien Huu Nguyen},
    title     = {Improving Aspect-based Sentiment Analysis with Gated Graph Convolutional Networks and Syntax-based Regulation},
    booktitle = {Findings of the Association for Computational Linguistics: EMNLP 2020},
    year      = {2020}
}

@inproceedings{feng2023ssemgat,
    author  = {Weize Quan and Yunfei Feng and Ming Zhou and Yunzhen Zhao and Tong Wang and Dong-Ming Yan},
    title   = {TCAN: Text-oriented Cross Attention Network for Multimodal Sentiment Analysis},
    booktitle = {arXiv:2404.04545},
    year    = {2024}
}

@inproceedings{yu2020gradient,
    author    = {Tianhe Yu and Saurabh Kumar and Abhishek Gupta and Sergey Levine and Karol Hausman and Chelsea Finn},
    title     = {Gradient Surgery for Multi-Task Learning},
    booktitle = {Advances in Neural Information Processing Systems (NeurIPS)},
    year      = {2020}
}

@inproceedings{liu2022metaweighting,
    author    = {Yuren Mao and Zekai Wang and Weiwei Liu and Xuemin Lin and Pengtao Xie},
    title     = {{MetaWeighting}: Learning to Weight Tasks in Multi-Task Learning},
    booktitle = {Findings of the Association for Computational Linguistics: ACL 2022},
    year      = {2022}
}

@article{touvron2024llama3,
    author  = {Aaron Grattafiori and Abhimanyu Dubey and Abhinav Jauhri and Abhinav Pandey and Abhishek Kadian and Ahmad Al-Dahle and Aiesha Letman and Akhil Mathur and Alan Schelten and Alex Vaughan and Amy Yang and Angela Fan and Anirudh Goyal and Anthony Hartshorn and Aobo Yang and Archi Mitra and Archie Sravankumar and Artem Korenev and Arthur Hinsvark and Arun Rao and Aston Zhang and Aurelien Rodriguez and Austen Gregerson and Ava Spataru and Baptiste Roziere and Bethany Biron and Binh Tang and Bobbie Chern and Charlotte Caucheteux and Chaya Nayak and Chloe Bi and Chris Marra and Chris McConnell and Christian Keller and Christophe Touret and Chunyang Wu and Corinne Wong and Cristian Canton Ferrer and Cyrus Nikolaidis and Damien Allonsius and Daniel Song and Danielle Pintz and Danny Livshits and Danny Wyatt and David Esiobu and Dhruv Choudhary and Dhruv Mahajan and Diego Garcia-Olano and Diego Perino and Dieuwke Hupkes and Egor Lakomkin and Ehab AlBadawy and Elina Lobanova and Emily Dinan and Eric Michael Smith and Filip Radenovic and Francisco Guzmán and Frank Zhang and Gabriel Synnaeve and Gabrielle Lee and Georgia Lewis Anderson and Govind Thattai and Graeme Nail and Gregoire Mialon and Guan Pang and Guillem Cucurell and Hailey Nguyen and Hannah Korevaar and Hu Xu and Hugo Touvron and Iliyan Zarov and Imanol Arrieta Ibarra and Isabel Kloumann and Ishan Misra and Ivan Evtimov and Jack Zhang and Jade Copet and Jaewon Lee and Jan Geffert and Jana Vranes and Jason Park and Jay Mahadeokar and Jeet Shah and Jelmer van der Linde and Jennifer Billock and Jenny Hong and Jenya Lee and Jeremy Fu and Jianfeng Chi and Jianyu Huang and Jiawen Liu and Jie Wang and Jiecao Yu and Joanna Bitton and Joe Spisak and Jongsoo Park and Joseph Rocca and Joshua Johnstun and Joshua Saxe and Junteng Jia and Kalyan Vasuden Alwala and Karthik Prasad and Kartikeya Upasani and Kate Plawiak and Ke Li and Kenneth Heafield and Kevin Stone and Khalid El-Arini and Krithika Iyer and Kshitiz Malik and Kuenley Chiu and Kunal Bhalla and Kushal Lakhotia and Lauren Rantala-Yeary and Laurens van der Maaten and Lawrence Chen and Liang Tan and Liz Jenkins and Louis Martin and Lovish Madaan and Lubo Malo and Lukas Blecher and Lukas Landzaat and Luke de Oliveira and Madeline Muzzi and Mahesh Pasupuleti and Mannat Singh and Manohar Paluri and Marcin Kardas and Maria Tsimpoukelli and Mathew Oldham and Mathieu Rita and Maya Pavlova and Melanie Kambadur and Mike Lewis and Min Si and Mitesh Kumar Singh and Mona Hassan and Naman Goyal and Narjes Torabi and Nikolay Bashlykov and Nikolay Bogoychev and Niladri Chatterji and Ning Zhang and Olivier Duchenne and Onur Çelebi and Patrick Alrassy and Pengchuan Zhang and Pengwei Li and Petar Vasic and Peter Weng and Prajjwal Bhargava and Pratik Dubal and Praveen Krishnan and Punit Singh Koura and Puxin Xu and Qing He and Qingxiao Dong and Ragavan Srinivasan and Raj Ganapathy and Ramon Calderer and Ricardo Silveira Cabral and Robert Stojnic and Roberta Raileanu and Rohan Maheswari and Rohit Girdhar and Rohit Patel and Romain Sauvestre and Ronnie Polidoro and Roshan Sumbaly and Ross Taylor and Ruan Silva and Rui Hou and Rui Wang and Saghar Hosseini and Sahana Chennabasappa and Sanjay Singh and Sean Bell and Seohyun Sonia Kim and Sergey Edunov and Shaoliang Nie and Sharan Narang and Sharath Raparthy and Sheng Shen and Shengye Wan and Shruti Bhosale and Shun Zhang and Simon Vandenhende and Soumya Batra and Spencer Whitman and Sten Sootla and Stephane Collot and Suchin Gururangan and Sydney Borodinsky and Tamar Herman and Tara Fowler and Tarek Sheasha and Thomas Georgiou and Thomas Scialom and Tobias Speckbacher and Todor Mihaylov and Tong Xiao and Ujjwal Karn and Vedanuj Goswami and Vibhor Gupta and Vignesh Ramanathan and Viktor Kerkez and Vincent Gonguet and Virginie Do and Vish Vogeti and Vítor Albiero and Vladan Petrovic and Weiwei Chu and Wenhan Xiong and Wenyin Fu and Whitney Meers and Xavier Martinet and Xiaodong Wang and Xiaofang Wang and Xiaoqing Ellen Tan and Xide Xia and Xinfeng Xie and Xuchao Jia and Xuewei Wang and Yaelle Goldschlag and Yashesh Gaur and Yasmine Babaei and Yi Wen and Yiwen Song and Yuchen Zhang and Yue Li and Yuning Mao and Zacharie Delpierre Coudert and Zheng Yan and Zhengxing Chen and Zoe Papakipos and Aaditya Singh and Aayushi Srivastava and Abha Jain and Adam Kelsey and Adam Shajnfeld and Adithya Gangidi and Adolfo Victoria and Ahuva Goldstand and Ajay Menon and Ajay Sharma and Alex Boesenberg and Alexei Baevski and Allie Feinstein and Amanda Kallet and Amit Sangani and Amos Teo and Anam Yunus and Andrei Lupu and Andres Alvarado and Andrew Caples and Andrew Gu and Andrew Ho and Andrew Poulton and Andrew Ryan and Ankit Ramchandani and Annie Dong and Annie Franco and Anuj Goyal and Aparajita Saraf and Arkabandhu Chowdhury and Ashley Gabriel and Ashwin Bharambe and Assaf Eisenman and Azadeh Yazdan and Beau James and Ben Maurer and Benjamin Leonhardi and Bernie Huang and Beth Loyd and Beto De Paola and Bhargavi Paranjape and Bing Liu and Bo Wu and Boyu Ni and Braden Hancock and Bram Wasti and Brandon Spence and Brani Stojkovic and Brian Gamido and Britt Montalvo and Carl Parker and Carly Burton and Catalina Mejia and Ce Liu and Changhan Wang and Changkyu Kim and Chao Zhou and Chester Hu and Ching-Hsiang Chu and Chris Cai and Chris Tindal and Christoph Feichtenhofer and Cynthia Gao and Damon Civin and Dana Beaty and Daniel Kreymer and Daniel Li and David Adkins and David Xu and Davide Testuggine and Delia David and Devi Parikh and Diana Liskovich and Didem Foss and Dingkang Wang and Duc Le and Dustin Holland and Edward Dowling and Eissa Jamil and Elaine Montgomery and Eleonora Presani and Emily Hahn and Emily Wood and Eric-Tuan Le and Erik Brinkman and Esteban Arcaute and Evan Dunbar and Evan Smothers and Fei Sun and Felix Kreuk and Feng Tian and Filippos Kokkinos and Firat Ozgenel and Francesco Caggioni and Frank Kanayet and Frank Seide and Gabriela Medina Florez and Gabriella Schwarz and Gada Badeer and Georgia Swee and Gil Halpern and Grant Herman and Grigory Sizov and Guangyi and Zhang and Guna Lakshminarayanan and Hakan Inan and Hamid Shojanazeri and Han Zou and Hannah Wang and Hanwen Zha and Haroun Habeeb and Harrison Rudolph and Helen Suk and Henry Aspegren and Hunter Goldman and Hongyuan Zhan and Ibrahim Damlaj and Igor Molybog and Igor Tufanov and Ilias Leontiadis and Irina-Elena Veliche and Itai Gat and Jake Weissman and James Geboski and James Kohli and Janice Lam and Japhet Asher and Jean-Baptiste Gaya and Jeff Marcus and Jeff Tang and Jennifer Chan and Jenny Zhen and Jeremy Reizenstein and Jeremy Teboul and Jessica Zhong and Jian Jin and Jingyi Yang and Joe Cummings and Jon Carvill and Jon Shepard and Jonathan McPhie and Jonathan Torres and Josh Ginsburg and Junjie Wang and Kai Wu and Kam Hou U and Karan Saxena and Kartikay Khandelwal and Katayoun Zand and Kathy Matosich and Kaushik Veeraraghavan and Kelly Michelena and Keqian Li and Kiran Jagadeesh and Kun Huang and Kunal Chawla and Kyle Huang and Lailin Chen and Lakshya Garg and Lavender A and Leandro Silva and Lee Bell and Lei Zhang and Liangpeng Guo and Licheng Yu and Liron Moshkovich and Luca Wehrstedt and Madian Khabsa and Manav Avalani and Manish Bhatt and Martynas Mankus and Matan Hasson and Matthew Lennie and Matthias Reso and Maxim Groshev and Maxim Naumov and Maya Lathi and Meghan Keneally and Miao Liu and Michael L. Seltzer and Michal Valko and Michelle Restrepo and Mihir Patel and Mik Vyatskov and Mikayel Samvelyan and Mike Clark and Mike Macey and Mike Wang and Miquel Jubert Hermoso and Mo Metanat and Mohammad Rastegari and Munish Bansal and Nandhini Santhanam and Natascha Parks and Natasha White and Navyata Bawa and Nayan Singhal and Nick Egebo and Nicolas Usunier and Nikhil Mehta and Nikolay Pavlovich Laptev and Ning Dong and Norman Cheng and Oleg Chernoguz and Olivia Hart and Omkar Salpekar and Ozlem Kalinli and Parkin Kent and Parth Parekh and Paul Saab and Pavan Balaji and Pedro Rittner and Philip Bontrager and Pierre Roux and Piotr Dollar and Polina Zvyagina and Prashant Ratanchandani and Pritish Yuvraj and Qian Liang and Rachad Alao and Rachel Rodriguez and Rafi Ayub and Raghotham Murthy and Raghu Nayani and Rahul Mitra and Rangaprabhu Parthasarathy and Raymond Li and Rebekkah Hogan and Robin Battey and Rocky Wang and Russ Howes and Ruty Rinott and Sachin Mehta and Sachin Siby and Sai Jayesh Bondu and Samyak Datta and Sara Chugh and Sara Hunt and Sargun Dhillon and Sasha Sidorov and Satadru Pan and Saurabh Mahajan and Saurabh Verma and Seiji Yamamoto and Sharadh Ramaswamy and Shaun Lindsay and Shaun Lindsay and Sheng Feng and Shenghao Lin and Shengxin Cindy Zha and Shishir Patil and Shiva Shankar and Shuqiang Zhang and Shuqiang Zhang and Sinong Wang and Sneha Agarwal and Soji Sajuyigbe and Soumith Chintala and Stephanie Max and Stephen Chen and Steve Kehoe and Steve Satterfield and Sudarshan Govindaprasad and Sumit Gupta and Summer Deng and Sungmin Cho and Sunny Virk and Suraj Subramanian and Sy Choudhury and Sydney Goldman and Tal Remez and Tamar Glaser and Tamara Best and Thilo Koehler and Thomas Robinson and Tianhe Li and Tianjun Zhang and Tim Matthews and Timothy Chou and Tzook Shaked and Varun Vontimitta and Victoria Ajayi and Victoria Montanez and Vijai Mohan and Vinay Satish Kumar and Vishal Mangla and Vlad Ionescu and Vlad Poenaru and Vlad Tiberiu Mihailescu and Vladimir Ivanov and Wei Li and Wenchen Wang and Wenwen Jiang and Wes Bouaziz and Will Constable and Xiaocheng Tang and Xiaojian Wu and Xiaolan Wang and Xilun Wu and Xinbo Gao and Yaniv Kleinman and Yanjun Chen and Ye Hu and Ye Jia and Ye Qi and Yenda Li and Yilin Zhang and Ying Zhang and Yossi Adi and Youngjin Nam and Yu and Wang and Yu Zhao and Yuchen Hao and Yundi Qian and Yunlu Li and Yuzi He and Zach Rait and Zachary DeVito and Zef Rosnbrick and Zhaoduo Wen and Zhenyu Yang and Zhiwei Zhao and Zhiyu Ma},
    title   = { title={The Llama 3 Herd of Models},},
    journal = {arXiv preprint arXiv:2407.21783},
    year    = {2024}
}

@article{liu2024deepseek,
    author  = {DeepSeek-AI and Aixin Liu and Bei Feng and Bing Xue and Bingxuan Wang and Bochao Wu and Chengda Lu and Chenggang Zhao and Chengqi Deng and Chenyu Zhang and Chong Ruan and Damai Dai and Daya Guo and Dejian Yang and Deli Chen and Dongjie Ji and Erhang Li and Fangyun Lin and Fucong Dai and Fuli Luo and Guangbo Hao and Guanting Chen and Guowei Li and H. Zhang and Han Bao and Hanwei Xu and Haocheng Wang and Haowei Zhang and Honghui Ding and Huajian Xin and Huazuo Gao and Hui Li and Hui Qu and J. L. Cai and Jian Liang and Jianzhong Guo and Jiaqi Ni and Jiashi Li and Jiawei Wang and Jin Chen and Jingchang Chen and Jingyang Yuan and Junjie Qiu and Junlong Li and Junxiao Song and Kai Dong and Kai Hu and Kaige Gao and Kang Guan and Kexin Huang and Kuai Yu and Lean Wang and Lecong Zhang and Lei Xu and Leyi Xia and Liang Zhao and Litong Wang and Liyue Zhang and Meng Li and Miaojun Wang and Mingchuan Zhang and Minghua Zhang and Minghui Tang and Mingming Li and Ning Tian and Panpan Huang and Peiyi Wang and Peng Zhang and Qiancheng Wang and Qihao Zhu and Qinyu Chen and Qiushi Du and R. J. Chen and R. L. Jin and Ruiqi Ge and Ruisong Zhang and Ruizhe Pan and Runji Wang and Runxin Xu and Ruoyu Zhang and Ruyi Chen and S. S. Li and Shanghao Lu and Shangyan Zhou and Shanhuang Chen and Shaoqing Wu and Shengfeng Ye and Shengfeng Ye and Shirong Ma and Shiyu Wang and Shuang Zhou and Shuiping Yu and Shunfeng Zhou and Shuting Pan and T. Wang and Tao Yun and Tian Pei and Tianyu Sun and W. L. Xiao and Wangding Zeng and Wanjia Zhao and Wei An and Wen Liu and Wenfeng Liang and Wenjun Gao and Wenqin Yu and Wentao Zhang and X. Q. Li and Xiangyue Jin and Xianzu Wang and Xiao Bi and Xiaodong Liu and Xiaohan Wang and Xiaojin Shen and Xiaokang Chen and Xiaokang Zhang and Xiaosha Chen and Xiaotao Nie and Xiaowen Sun and Xiaoxiang Wang and Xin Cheng and Xin Liu and Xin Xie and Xingchao Liu and Xingkai Yu and Xinnan Song and Xinxia Shan and Xinyi Zhou and Xinyu Yang and Xinyuan Li and Xuecheng Su and Xuheng Lin and Y. K. Li and Y. Q. Wang and Y. X. Wei and Y. X. Zhu and Yang Zhang and Yanhong Xu and Yanhong Xu and Yanping Huang and Yao Li and Yao Zhao and Yaofeng Sun and Yaohui Li and Yaohui Wang and Yi Yu and Yi Zheng and Yichao Zhang and Yifan Shi and Yiliang Xiong and Ying He and Ying Tang and Yishi Piao and Yisong Wang and Yixuan Tan and Yiyang Ma and Yiyuan Liu and Yongqiang Guo and Yu Wu and Yuan Ou and Yuchen Zhu and Yuduan Wang and Yue Gong and Yuheng Zou and Yujia He and Yukun Zha and Yunfan Xiong and Yunxian Ma and Yuting Yan and Yuxiang Luo and Yuxiang You and Yuxuan Liu and Yuyang Zhou and Z. F. Wu and Z. Z. Ren and Zehui Ren and Zhangli Sha and Zhe Fu and Zhean Xu and Zhen Huang and Zhen Zhang and Zhenda Xie and Zhengyan Zhang and Zhewen Hao and Zhibin Gou and Zhicheng Ma and Zhigang Yan and Zhihong Shao and Zhipeng Xu and Zhiyu Wu and Zhongyu Zhang and Zhuoshu Li and Zihui Gu and Zijia Zhu and Zijun Liu and Zilin Li and Ziwei Xie and Ziyang Song and Ziyi Gao and Zizheng Pan},
    title   = {DeepSeek-V3 Technical Report},
    journal = {arXiv preprint arXiv:2412.19437},
    year    = {2024}
}

@article{yang2024qwen2,
    author  = {Qwen and : and An Yang and Baosong Yang and Beichen Zhang and Binyuan Hui and Bo Zheng and Bowen Yu and Chengyuan Li and Dayiheng Liu and Fei Huang and Haoran Wei and Huan Lin and Jian Yang and Jianhong Tu and Jianwei Zhang and Jianxin Yang and Jiaxi Yang and Jingren Zhou and Junyang Lin and Kai Dang and Keming Lu and Keqin Bao and Kexin Yang and Le Yu and Mei Li and Mingfeng Xue and Pei Zhang and Qin Zhu and Rui Men and Runji Lin and Tianhao Li and Tianyi Tang and Tingyu Xia and Xingzhang Ren and Xuancheng Ren and Yang Fan and Yang Su and Yichang Zhang and Yu Wan and Yuqiong Liu and Zeyu Cui and Zhenru Zhang and Zihan Qiu},
    title   = {Qwen2.5 Technical Report},
    journal = {arXiv preprint arXiv:2412.15115},
    year    = {2024}
}

@article{schulman2025lora,
  author = {John Schulman and Thinking Machines Lab},
  title = {LoRA Without Regret},
  journal = {Thinking Machines Lab: Connectionism},
  year = {2025},
  note = {https://thinkingmachines.ai/blog/lora/},
  doi = {10.64434/tml.20250929},
}

@inproceedings{su2017deep,
    author    = {Yanzhu Liu and Adams Wai Kin Kong and Chi Keong Goh},
    title     = {Deep Ordinal Regression Based on Data Relationship for Small Datasets},
    booktitle = {Proceedings of the 26th International Joint Conference on Artificial Intelligence (IJCAI)},
    year      = {2017}
}

@article{kook2022deep,
    author  = {Lucas Kook and Lisa Herzog and Torsten Hothorn and Oliver Dürr and Beate Sick},
    title   = {Deep and Interpretable Regression Models for Ordinal Outcomes},
    journal = {Pattern Recognition},
    volume  = {122},
    pages   = {108263},
    year    = {2022}
}

@inproceedings{cheryfs2023contrastive,
    author    = {Weiwei Wang and Wenyang Wei and Qingyuan Song and Yansong Wang},
    title     = {Leveraging Contrastive Learning with BERT for ESG Issue Identification},
    booktitle = {Proceedings of the Fifth Workshop on Financial Technology and Natural Language Processing and the Second Multimodal AI For Financial Forecasting},
    year      = {2023}
}

@article{ni2024peft,
    author  = {Zihao Fu and Haoran Yang and Anthony Man-Cho So and Wai Lam and Lidong Bing and Nigel Collier},
    title   = {On the Effectiveness of Parameter-Efficient Fine-Tuning},
    journal = {arXiv:2211.15583},
    year    = {2022}
}

@inproceedings{zhou2024autopeft,
    author    = {Han Zhou and Xingchen Wan and Ivan Vulić and Anna Korhonen},
    title     = {{AutoPEFT}: Automatic Configuration Search for Parameter-Efficient Fine-Tuning},
    booktitle = {Transactions of the Association for Computational Linguistics, Volume 12},
    year      = {2024}
}

@inproceedings{ploner2024selective,
    author = {Max Ploner and Alan Akbik},
    title = {Parameter-Efficient Fine-Tuning: Is There An Optimal Subset of Parameters to Tune ?},
    booktitle = {Findings of the Association for Computational Linguistics: EACL 2024},
    year = {2024},
}

@article{vuckovic2023kfold,
    author  = {Juan M Gorriz and R. Martin Clemente and F Segovia and J Ramirez and A Ortiz and J. Suckling},
    title   = {Is K-fold cross validation the best model selection method for Machine Learning ?},
    journal = {arXiv:2401.16407},
    year    = {2024}
}

@article{nomad2023lora,
    author  = {Minghao Yan and Zhuang Wang and Zhen Jia and Shivaram Venkataraman and Yida Wang},
    title   = {PLoRA: Efficient LoRA Hyperparameter Tuning for Large Models},
    journal = {arXiv:2508.02932},
    year    = {2025}
}

@inproceedings{takamoto2025temperature,
    author    = {Bum Jun Kim and Sang Woo Kim},
    title     = {Temperature-Free Loss Function for Contrastive Learning},
    booktitle = {arXiv:2501.17683},
    year      = {2025}
}

@inproceedings{becerra2024ordinal,
    author    = {Daehwan Kim and Haejun Chung and Ikbeom Jang},
    title     = {Calibration of ordinal regression networks},
    booktitle = {arXiv:2410.15658},
    year      = {2024}
}

@inproceedings{qiu2025gated,
    author    = {Zihan Qiu and Zekun Wang and Bo Zheng and Zeyu Huang and Kaiyue Wen and Songlin Yang and Rui Men and Le Yu and Fei Huang and Suozhi Huang and Dayiheng Liu and Jingren Zhou and Junyang Lin},
    title     = {Gated Attention for Large Language Models: Non-linearity, Sparsity, and Stability},
    booktitle = {Advances in Neural Information Processing Systems (NeurIPS)},
    year      = {2025},
}

@inproceedings{kumar2025llm,
    author    = {Alex Ning and Vainateya Rangaraju},
    title     = {Visualizing LLM Latent Space Geometry Through Dimensionality Reduction},
    booktitle = {arXiv preprint arXiv:2511.21594},
    year      = {2025},
}

@article{kobak2023umap,
    author  = {Dmitry Kobak and Philipp Berens},
    title   = {The Art of Using {t-SNE} and {UMAP} for Single-Cell Transcriptomics},
    journal = {Nature Communications},
    volume  = {10},
    pages   = {5416},
    year    = {2023}
}

@inproceedings{senties2021umap,
    author={Alex Ning and Vainateya Rangaraju},
    title={Visualizing LLM Latent Space Geometry Through Dimensionality Reduction},
    booktitle={arXiv preprint arXiv:2511.21594},
    year={2025},
}

@inproceedings{vujanic2025umap,
    author={Mohammad Tariqul Islam and Jason W. Fleischer},
    title={The Shape of Attraction in UMAP: Exploring the Embedding Forces in Dimensionality Reduction},
    booktitle={arXiv preprint arXiv:2503.09101},
    year={2025},
}

@inproceedings{zhang2024cl,
    author    = {Jeff Calder and Wonjun Lee},
    title     = {Understanding Contrastive Learning through Variational Analysis and Neural Network Optimization Perspectives},
    booktitle   = {arXiv preprint arXiv:2503.10812},
    year      = {2025},
    url       = {https://arxiv.org/abs/2503.10812}
}

@article{gronbech2021scrna,
    author  = {Christopher Heje Grønbech and Maximillian Fornitz Vording and Pascal N Timshel and Casper Kaae Sønderby and Tune H Pers and Ole Winther},
    title   = {{scVAE}: Variational Auto-Encoders for Single-Cell Gene Expression Data},
    journal = {Bioinformatics},
    volume  = {36},
    number  = {16},
    pages   = {4415--4422},
    year    = {2020}
}

@inproceedings{zela2023nomad,
    author    = {Christophe Tribes and Sacha Benarroch-Lelong and Peng Lu and Ivan Kobyzev},
    title     = {Hyperparameter Optimization for Large Language Model Instruction-Tuning},
    booktitle = {arXiv:2312.00949},
    year      = {2023}
}

@inproceedings{xu2025lorapro,
    author    = {Zhengbo Wang and Jian Liang and Ran He and Zilei Wang and Tieniu Tan},
    title     = {{LoRA-Pro}: Are Low-Rank Adapters Properly Optimized ?},
    booktitle = {Proceedings of the 13th International Conference on Learning Representations (ICLR)},
    year      = {2025}
}

@inproceedings{li2025trainsmall,
    author    = {Jun Zhang and Jue Wang and Huan Li and Lidan Shou and Ke Chen and Yang You and Guiming Xie and Xuejian Gong and Kunlong Zhou},
    title     = {Train Small, Infer Large: Memory-Efficient LoRA Training for Large Language Models},
    booktitle = {arxiv:2502.13533},
    year      = {2025}
}

@inproceedings{jeong2024angle,
    author    = {Yoo Hyun Jeong and Myeongsoo Han and Dong-Kyu Chae},
    title     = {A Simple Angle-based Approach for Contrastive Learning of Unsupervised Sentence Representation},
    booktitle = {Findings of the Association for Computational Linguistics: EMNLP 2024},
    year      = {2024},
}

@article{gatera2024token,
    author  = {Jie Ou and Shuaihong Jiang and Yingjun Du and Cees G. M. Snoek},
    title   = {GateRA: Token-Aware Modulation for Parameter-Efficient Fine-Tuning},
    journal = {arXiv:2511.17582 },
    year    = {2025}
}

@article{m2o2025multi,
    author  = {Matthew Barker and Andrew Bell and Evan Thomas and James Carr and Thomas Andrews and Umang Bhatt},
    title   = {Faster, Cheaper, Better: Multi-Objective Hyperparameter Optimization for LLM and RAG Systems},
    journal = {	arXiv:2502.18635},
    year    = {2025}
}

@inproceedings{gradmultinorm2025alternating,
    author    = {Meyer Scetbon and Chao Ma and Wenbo Gong and Edward Meeds},
    title     = {Gradient Multi-Normalization for Stateless and Scalable {LLM} Training},
    booktitle = {Advances in Neural Information Processing Systems (NeurIPS)},
    year      = {2025}
}

\newpage
\appendix

\section{A3CG Dataset Overview and Task Formulation}
\label{app:a3cg}

\subsection{Dataset Structure}

A3CG is a benchmark for aspect--action analysis of sustainability claims, where each claim may reference multiple ESG aspects and associated actions.
Each claim is annotated with:
\begin{enumerate}[nosep,leftmargin=*]
    \item one or more \textit{aspects},
    \item a higher-level \textit{aspect category} grouping related aspects, and
    \item an \textit{action label} capturing the degree of concreteness and commitment expressed in the claim.
\end{enumerate}

\subsection{Ordinal Action Labels}

Actions are organized along an ordinal scale reflecting increasing actionability and evidential grounding:
\begin{itemize}[nosep,leftmargin=*]
    \item \textbf{Indeterminate}: aspirational statements without concrete steps.
    \item \textbf{Planning}: future-oriented commitments or partial roadmaps.
    \item \textbf{Implemented}: actions described as carried out with verifiable details.
\end{itemize}

\subsection{Cross-Category Generalization Setting}

A3CG evaluates cross-category generalization by holding out a subset of aspect categories that are never observed during training or validation and are only evaluated at test time. Following prior work, we adopt a three-fold evaluation protocol, where each fold excludes a distinct set of aspect categories to form the unseen test split.

Since aspects from different categories frequently co-occur within the same claim, this setting differs from standard leave-one-domain-out evaluations. The protocol is therefore designed to minimize data leakage while preserving realistic statement composition, encouraging models to rely on transferable, action-centered representations rather than category-specific lexical cues.

\subsection{Relevance to Structured Representation Learning}

The ordinal structure of action labels and the cross-category evaluation protocol make A3CG well suited for studying structured representation learning.
In this work, A3CG serves as a testbed for assessing whether contrastive and ordinal supervision can shape latent representations that generalize beyond seen sustainability categories.
Table~\ref{tab:a3cg_examples} provides representative examples illustrating the aspect--category--action annotation structure.

\section{Pair Construction Procedures}
\label{app:pairs}

This appendix provides an algorithmic view of the pair construction procedures defined in Section~4 (Experimental Setup), translating Eqs.~\ref{eq:pos_neg_sets}--\ref{eq:transition_function} into reproducible steps and clarifying implementation details.

\subsection{Contrastive Pair Construction (Algorithmic Procedure)}
\label{app:contrastive_pairs}

Algorithm~\ref{alg:contrastive_pairs} implements the contrastive sampling rule in Eq.~\ref{eq:pos_neg_sets}. For each anchor claim $x_i$, positives are all candidates whose label sets intersect with $\mathcal{L}_i$, and negatives are all remaining candidates. We exclude self-pairs ($j=i$) and construct multi-positive sets for the loss in Section~3.

\begin{algorithm}[H]
\caption{Contrastive pair construction (multi-positive)}
\label{alg:contrastive_pairs}
\small
\begin{algorithmic}[1]
\REQUIRE Claims $\{x_i\}_{i=1}^{N}$ with label sets $\{\mathcal{L}_i\}$ (aspect- or category-level)
\FOR{$i=1$ \TO $N$}
  \STATE $\mathcal{P}(i)\leftarrow \emptyset,\;\mathcal{N}(i)\leftarrow \emptyset$
  \FOR{$j=1$ \TO $N$}
    \IF{$j=i$}
      \STATE \textbf{continue}
    \ENDIF
    \STATE $I_{ij}\leftarrow \mathcal{L}_i \cap \mathcal{L}_j$
    \IF{$|I_{ij}|>0$}
      \STATE $\mathcal{P}(i)\leftarrow \mathcal{P}(i)\cup\{x_j\}$
    \ELSE
      \STATE $\mathcal{N}(i)\leftarrow \mathcal{N}(i)\cup\{x_j\}$
    \ENDIF
  \ENDFOR
\ENDFOR
\end{algorithmic}
\end{algorithm}

\subsection{Ordinal Pair Construction (Algorithmic Procedure)}
\label{app:ordinal_pairs}

Algorithm~\ref{alg:ordinal_pairs} implements ordinal sampling using the transition function $\pi(\cdot)$ from Eq.~\ref{eq:transition_function}. For each anchor $x_i$, we first map its label set $\mathcal{L}_i$ into a dictionary $d_i:k\mapsto a$ (and similarly for $x_j$), where $k$ denotes an aspect or category key. A candidate is an ordinal positive if it shares at least one key with the anchor and matches the expected transition for some shared key; otherwise it is assigned as a negative (including the case of no shared keys).

\begin{algorithm}[H]
\caption{Ordinal pair construction (transition-based positives)}
\label{alg:ordinal_pairs}
\small
\begin{algorithmic}[1]
\REQUIRE Claims $\{x_i\}_{i=1}^{N}$ with label sets $\{\mathcal{L}_i\}$; transition $\pi(\cdot)$ (Eq.~\ref{eq:transition_function})
\FOR{$i=1$ \TO $N$}
  \STATE $d_i \leftarrow \textsc{Dict}(\mathcal{L}_i)$ \COMMENT{$k \mapsto a$}
  \STATE $\mathcal{P}_{\text{ord}}(i)\leftarrow \emptyset,\;\mathcal{N}_{\text{ord}}(i)\leftarrow \emptyset$
  \FOR{$j=1$ \TO $N$}
    \IF{$j=i$}
      \STATE \textbf{continue}
    \ENDIF
    \STATE $d_j \leftarrow \textsc{Dict}(\mathcal{L}_j)$
    \STATE $\mathcal{K}_{ij}\leftarrow \mathrm{keys}(d_i)\cap \mathrm{keys}(d_j)$
    \IF{$\mathcal{K}_{ij}=\emptyset$}
      \STATE $\mathcal{N}_{\text{ord}}(i)\leftarrow \mathcal{N}_{\text{ord}}(i)\cup\{x_j\}$
      \STATE \textbf{continue}
    \ENDIF
    \STATE $\textit{isPos}\leftarrow \textbf{false}$
    \FORALL{$k\in \mathcal{K}_{ij}$}
      \IF{$d_j(k)=\pi(d_i(k))$}
        \STATE $\textit{isPos}\leftarrow \textbf{true}$
        \STATE \textbf{break}
      \ENDIF
    \ENDFOR
    \IF{\textit{isPos}}
      \STATE $\mathcal{P}_{\text{ord}}(i)\leftarrow \mathcal{P}_{\text{ord}}(i)\cup\{x_j\}$
    \ELSE
      \STATE $\mathcal{N}_{\text{ord}}(i)\leftarrow \mathcal{N}_{\text{ord}}(i)\cup\{x_j\}$
    \ENDIF
  \ENDFOR
\ENDFOR
\end{algorithmic}
\end{algorithm}

\subsection{Aspect-Level vs.\ Category-Level Instantiation}
\label{app:granularity}

The procedures above are identical for aspect-level and category-level construction; only the definition of the key space differs. In aspect-level mode, keys $k$ correspond to fine-grained aspects. In category-level mode, aspects are first mapped to their parent categories, and keys correspond to categories. This mapping increases positive coverage (more shared keys per anchor) and encourages higher-level alignment across semantically related aspects, while preserving the same intersection and transition criteria.

\subsection{Implementation Notes}
\label{app:impl_notes}

Na\"ively constructing pairs is $O(N^2)$ in the number of claims. In practice, we precompute dictionaries $\{d_i\}$ once, cache key sets $\mathrm{keys}(d_i)$, and exclude self-pairs. All steps are deterministic given the dataset and the chosen granularity mode.

\section{Full Experimental Results}
\label{app:full_results}

This appendix reports the complete set of experimental results for all evaluated models, folds, and training configurations. While the main paper focuses on representative trends and best-performing configurations, the tables below provide exhaustive results to ensure full transparency and reproducibility.

\subsection{Evaluation Protocol Recap}

All results are reported in terms of F1 score for the Aspect--Action Analysis (AAA) task, evaluated separately on seen (S) and unseen (US) categories under the three-fold cross-category generalization protocol described in Section~4. For each configuration, we report results on all three folds as well as the full dataset, using the same format as Table~1 in the main paper.

\subsection{T5 Configurations}

For T5, we evaluate full fine-tuning as well as progressively structured configurations. The baseline fine-tuning setup uses a learning rate of $3\times10^{-5}$ with a batch size of 8.  

For contrastive learning, we perform a dedicated representation pre-training stage of up to 20 epochs with early stopping (patience = 3), using a learning rate of $1\times10^{-4}$, temperature $\tau=0.07$, and a maximum of 2 positive and 2 negative samples per anchor with a batch size of 6.  

Subsequent fine-tuning is conducted for 100 epochs, with the model checkpoint achieving the best validation performance selected for evaluation on the test set. For structured configurations beyond contrastive learning, we retain the same contrastive setup and incrementally introduce additional components. The ordinal loss is first added with a fixed margin of 0.05. For T5 specifically, gating and scaling are then applied simultaneously with $\lambda_c = 1.0$, $\lambda_{\text{o}} = 1.4$, $T_c = 5.8$, and $T_o = 1.0$. When MetaGradNorm is enabled, we use $\gamma = 0.5$, $\beta = 0.01$, and a meta-learning rate $\eta_{\text{meta}} = 0.001$ to adapt loss weights during training.

Table~\ref{tab:t5} reports full results for all T5 configurations across folds while Table~\ref{tab:t5_results_bis} provides a direct comparison between the COGLM framework and previous methods utilized in A3CG.

\subsection{Decoder-Only LLM Configurations}

For decoder-only models (LLaMA-3-8B, Gemma-7B, Mistral-7B, DeepSeek-7B, and Qwen2.5-7B), we employ parameter-efficient fine-tuning with LoRA. Unless otherwise specified, all models use a LoRA rank of $r=8$, scaling factor $\alpha=16$, and LoRA dropout of 0.05.

Contrastive learning is performed with a batch size of 5, a maximum of 3 positive and 6 negative samples per anchor, and 2 epochs of contrastive training. The contrastive learning rate is set to $1\times10^{-4}$.  

Subsequent LoRA fine-tuning is carried out for a minimum of 4 and a maximum of 6 epochs, using a batch size of 3 and a learning rate of $3\times10^{-5}$. For progressively structured configurations, components are introduced in stages while keeping previously selected hyperparameters fixed. Gated feature modulation is first applied with temperatures $T_c = 13$ and $T_o = 1$, followed by loss scaling with $\lambda_c = 1$ and $\lambda_{\text{o}} = 2.5$. MetaGradNorm is then added with $\gamma = 0.5$, $\beta = 0.01$, and a meta-learning rate $\eta_{\text{meta}} = 0.001$. The temperature and scaling parameters are selected empirically based on the initial magnitude ratio between contrastive and ordinal losses, ensuring balanced gradient contributions during training.

Tables~\ref{tab:llama-lora-8-16}--\ref{tab:qwen-lora-8-16} provide exhaustive results for all decoder-only models across folds and configurations.
Table~\ref{tab:llm_results_bis} provides a direct comparison between the COGLM framework applied on LLMs and previous methods utilized with LLMs in A3CG.

\subsection{Table Organization}

Each table follows the same layout as Table~1 in the main paper, reporting:
(i) the training configuration
and (ii) fold-wise Seen and Unseen F1 scores.  
This organization enables direct comparison across models, folds, and methodological components, and allows the reader to trace performance variations back to specific architectural or optimization choices.

\section{Hyperparameter Study -- LLaMA-3-8B}
\label{app:llama-hyperparams}

The staged exploration for LLaMA-3-8B is conducted according to the configurations detailed in Table~\ref{tab:llama-ablation}. Initial LoRA baseline experiments vary rank and scaling pairs $(r, \alpha) \in \{(2,4), (8,16), (16,32), (32,64)\}$ with corresponding fine-tuning learning rates $\eta_{\text{FT}} \in \{6 \times 10^{-5}, 3 \times 10^{-5}, 2 \times 10^{-5}, 1.5 \times 10^{-5}\}$. Across all configurations, models undergo 2 epochs of contrastive pre-training followed by 4 epochs of supervised fine-tuning.

For the contrastive stage, the learning rate $\eta_{\text{contrastive}}$ is tested across $\{10^{-3}, 10^{-4}, 5 \times 10^{-5}\}$ while fixing $\eta_{\text{FT}} = 3 \times 10^{-5}$. Subsequent two-loss training introduces an ordinal margin $m_0 \in \{0.05, 0.10, 0.15\}$. Gated configurations utilize temperature pairs $(T_c, T_o) \in \{(5,4), (13,1), (25,0.5)\}$, followed by loss scaling coefficients $(\lambda_c, \lambda_o) \in \{(2,1), (1,2.5), (0.5,5)\}$. Finally, MetaGradNorm stabilization is applied with parameters $\gamma \in \{0.2, 0.5, 1.0\}$, $\beta \in \{0.005, 0.01, 0.02\}$, and $\eta_{\text{meta}} \in \{3 \times 10^{-4}, 10^{-3}, 3 \times 10^{-3}\}$.

\section{Additional Visualizations}
\label{app:additional_visuals}

This appendix provides supplementary visualizations to extend the coverage of the analyses presented in Section~5. These figures are included for completeness and transparency, and do not introduce additional interpretations beyond those discussed in the main text.

\subsection{Seen--Unseen Performance Trajectories Across All Folds}

\begin{figure*}[t]
\centering
\includegraphics[width=\textwidth]{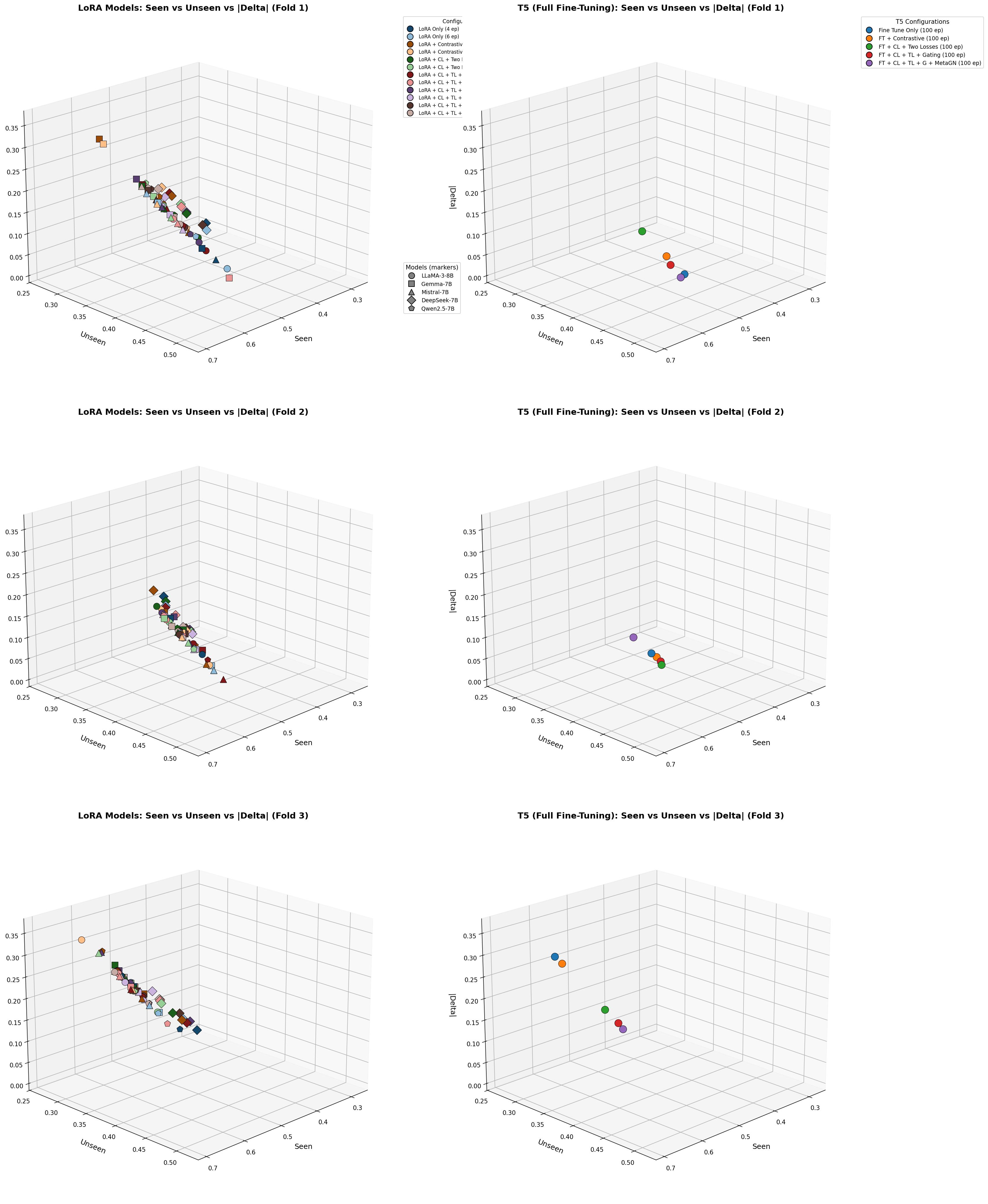}
\caption{Seen vs.\ Unseen vs.\ $|\Delta|$ across all folds. Left: LoRA-based LLMs; Right: T5 full fine-tuning. Colors denote configurations; markers denote model backbones.}
\label{fig:all_folds_3d_appendix}
\end{figure*}

Figure~\ref{fig:all_folds_3d_appendix} extends the three-dimensional visualization of seen performance, unseen performance, and $|\Delta|$ to all folds and model families. The figure reports trajectories for LoRA-based models and T5 full fine-tuning under identical plotting conventions, allowing inspection of fold-specific variability and configuration dispersion.

\subsection{Additional PCA and UMAP Projections}

\begin{figure*}[t]
\centering

\begin{minipage}{\textwidth}
    \centering
    \includegraphics[width=0.75\textwidth]{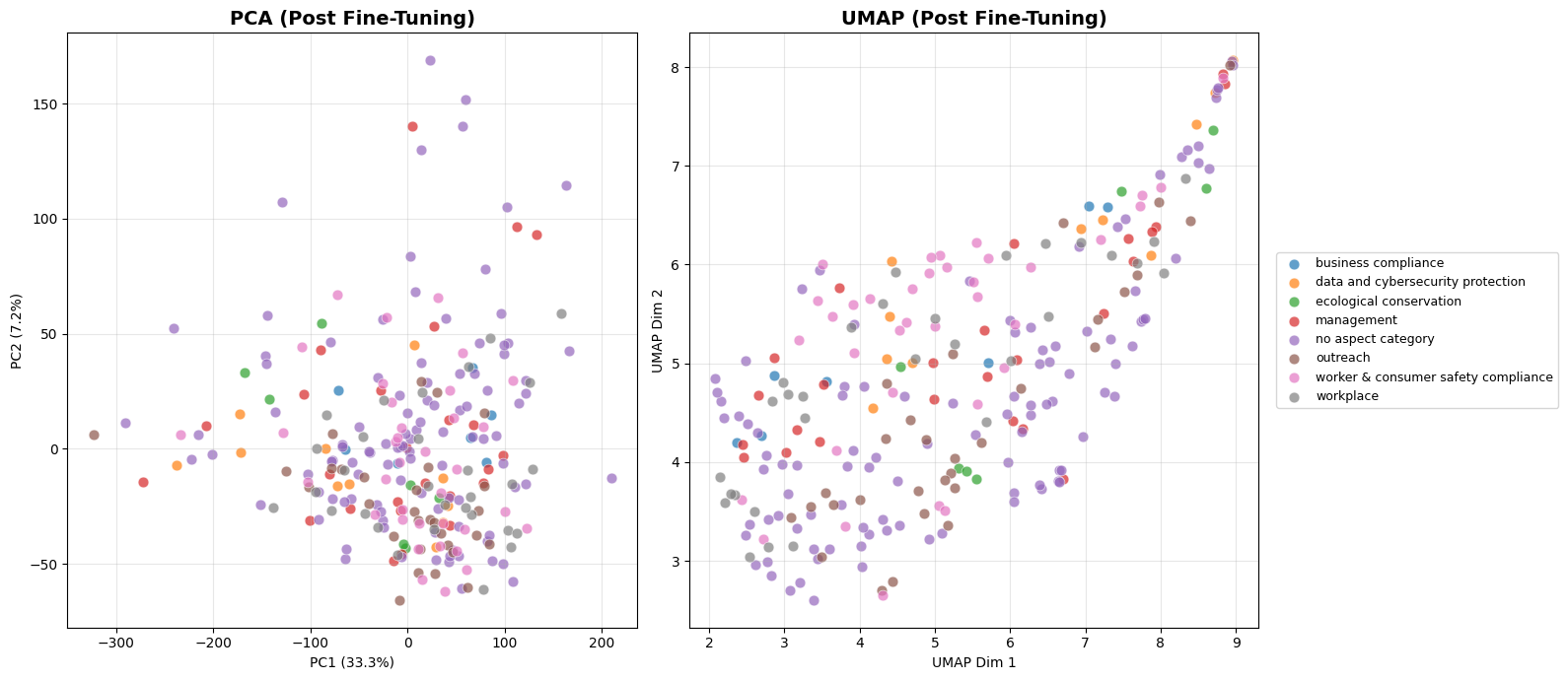}\\[-0.3em]
    {\scriptsize Gemma-7B: LoRA-only baseline | Sil.: $-0.067$ | CH: $2.47$ | Sep.: $0.37$}
\end{minipage}

\vspace{0.3em}

\begin{minipage}{\textwidth}
    \centering
    \includegraphics[width=0.75\textwidth]{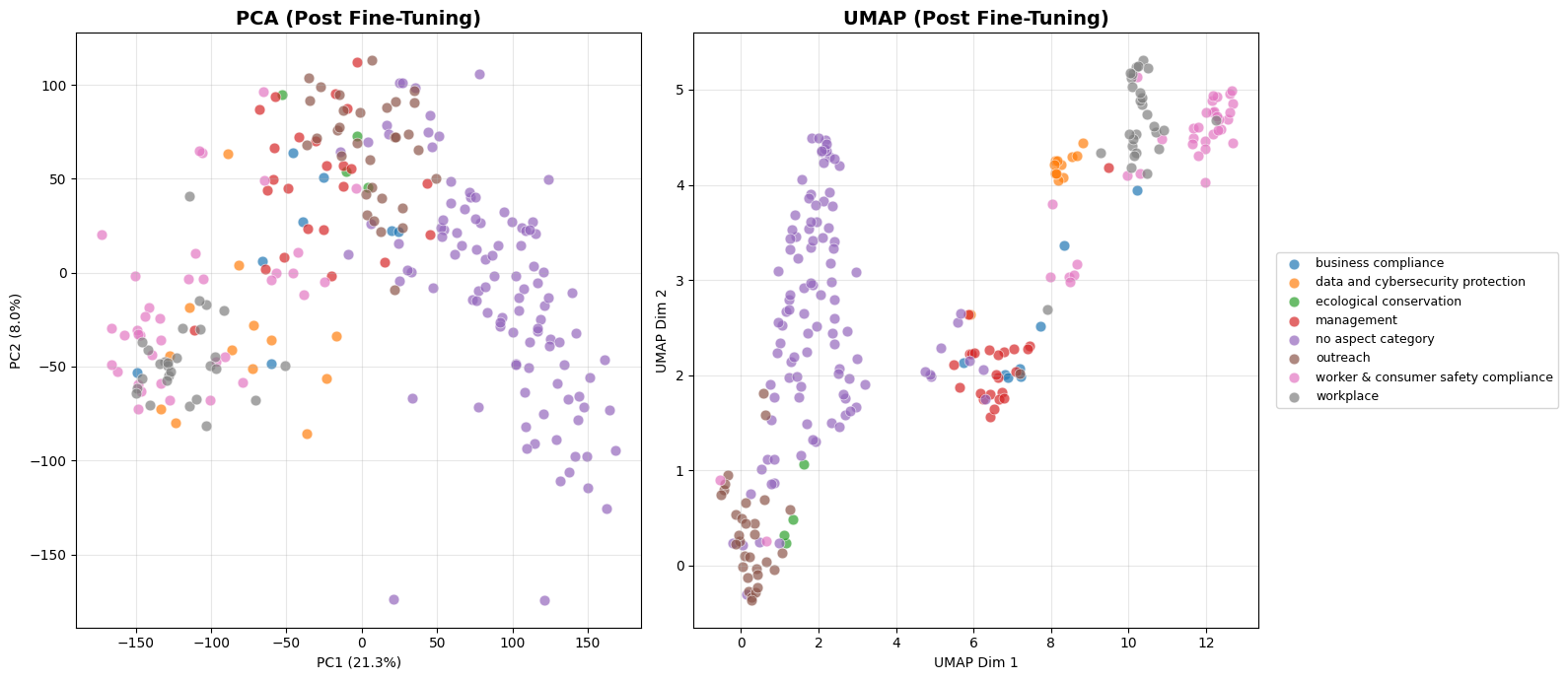}\\[-0.3em]
    {\scriptsize Gemma-7B: Full structured framework | Sil.: $0.082$ | CH: $16.67$ | Sep.: $0.79$}
\end{minipage}

\vspace{0.6em}
\noindent\rule{\textwidth}{0.4pt}
\vspace{0.4em}

\begin{minipage}{\textwidth}
    \centering
    \includegraphics[width=0.75\textwidth]{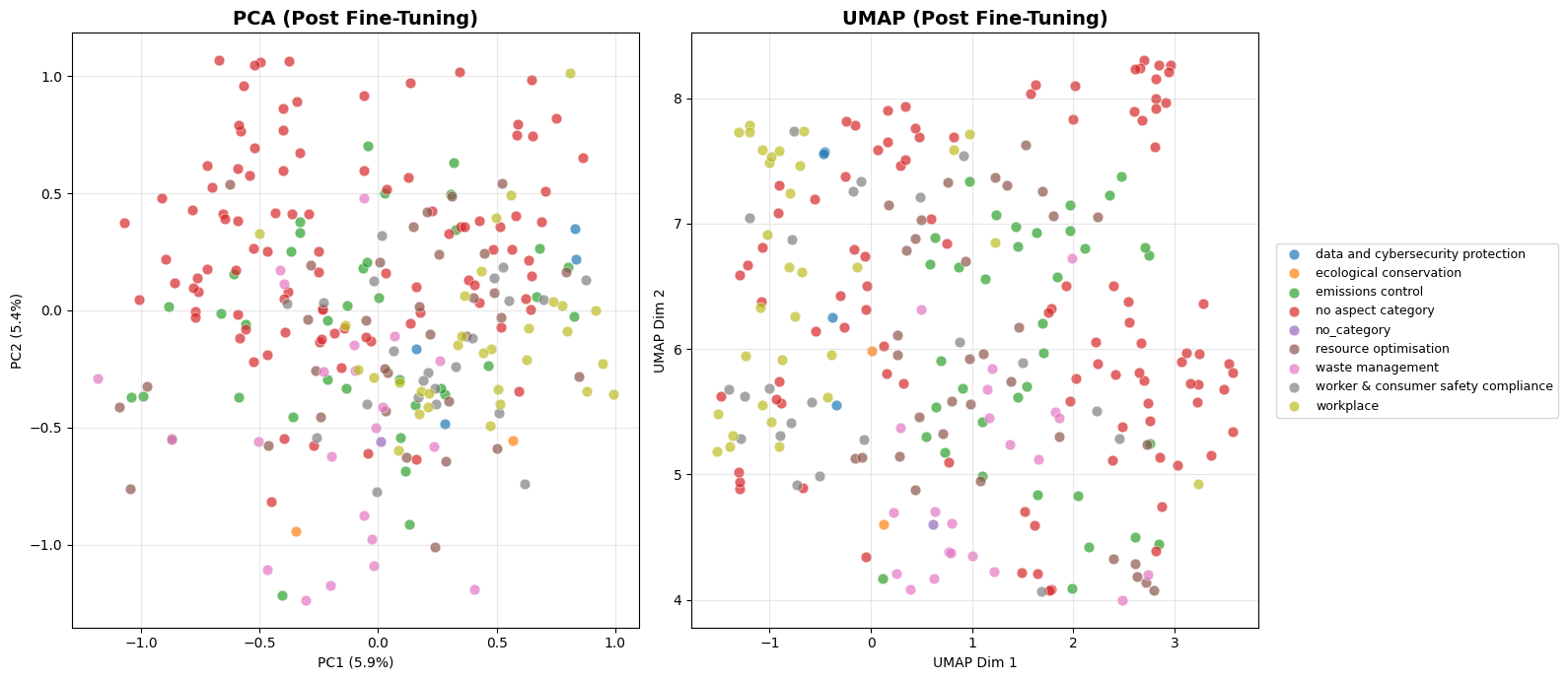}\\[-0.3em]
    {\scriptsize T5: Fine-tuning-only baseline | Sil.: $-0.030$ | CH: $3.27$ | Sep.: $0.52$}
\end{minipage}

\vspace{0.3em}

\begin{minipage}{\textwidth}
    \centering
    \includegraphics[width=0.75\textwidth]{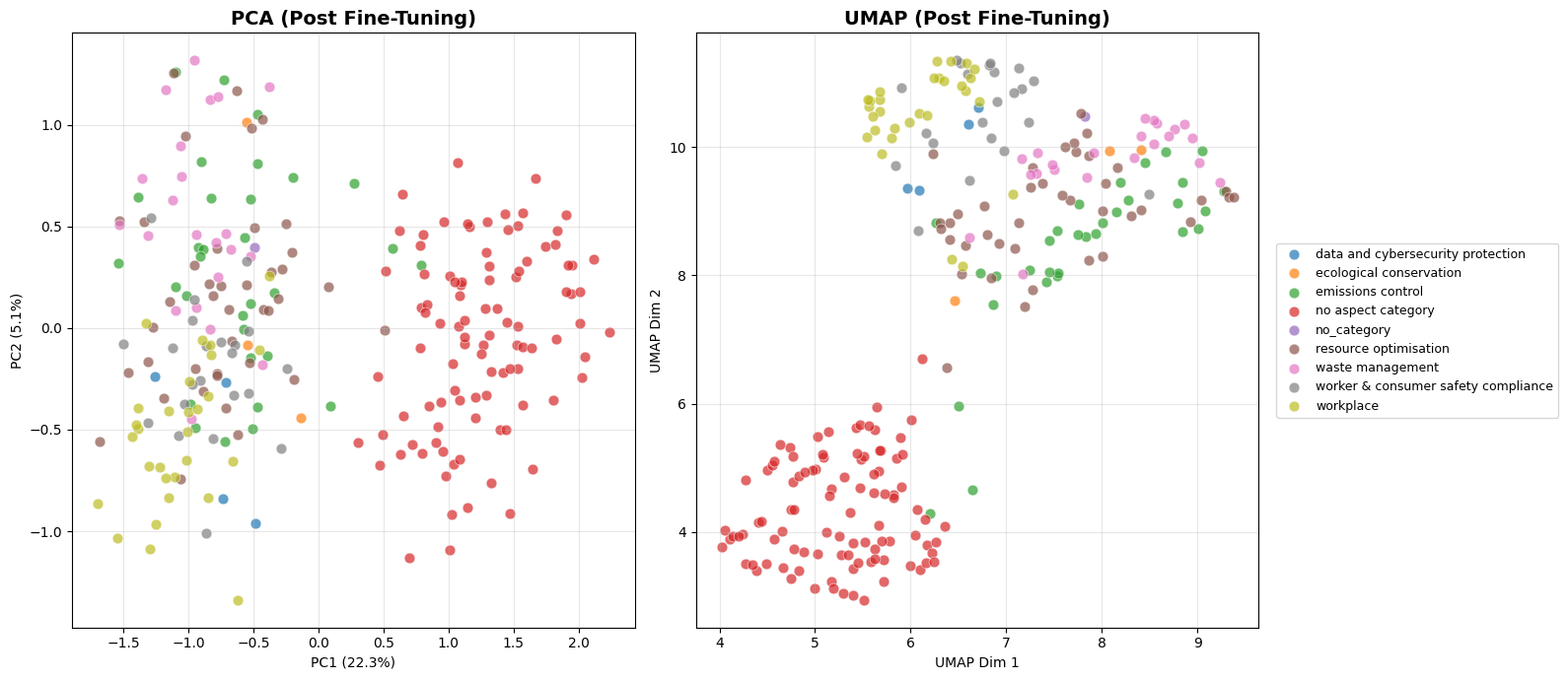}\\[-0.3em]
    {\scriptsize T5: Full structured framework | Sil.: $0.047$ | CH: $12.00$ | Sep.: $0.62$}
\end{minipage}

\caption{PCA and UMAP visualizations for Gemma-7B (Fold 1, top) and T5 (Fold 3, bottom). Baselines show diffuse clusters; full frameworks yield tighter separation. Sil.=Silhouette, CH=Calinski-Harabasz, Sep.=Separation Ratio.}
\label{fig:embedding_comparison}
\end{figure*}

Figure~\ref{fig:embedding_comparison} presents PCA and UMAP projections of learned representations for Gemma-7B (Fold~1) and T5 (Fold~3). For Gemma-7B, we compare the LoRA-only baseline against LoRA + CL + TL + G, the second-best configuration on Fold~1, selected to illustrate embedding structure under a different model than T5. For T5, we compare the fine-tuning-only baseline against FT + CL + TL + G + L + MGN, the best-performing configuration on Fold~3.

\section{Implementation and Training Details}
\label{app:implementation}

All experiments were conducted using publicly available frameworks, including PyTorch and the HuggingFace Transformers and PEFT libraries. Training was performed on NVIDIA A100 GPUs with 80,GB of memory, using mixed-precision (FP16) to reduce computational overhead. Optimization was carried out with AdamW, and reproducibility was ensured by fixing the random seed to 155 for Python, NumPy, and PyTorch. Training times were typically under one hour for T5 fine-tuning and approximately two hours for decoder-only LLM configurations, with early stopping applied during contrastive pre-training stages and validation-based checkpoint selection used for final evaluation.

\begin{table*}[!t]
\centering
\small
\begin{tabularx}{\textwidth}{Xlll}
\toprule
\textbf{Claim (excerpt)} & \textbf{Aspect} & \textbf{Category} & \textbf{Action} \\
\midrule
``We aim to reduce our carbon footprint over the coming years.'' 
& Emissions reduction & Environment & Indeterminate \\
\addlinespace[2pt]
``The company plans to invest in renewable energy infrastructure by 2026.'' 
& Energy transition & Environment & Planning \\
\addlinespace[2pt]
``In 2023, we installed solar panels across all manufacturing sites.'' 
& Renewable energy & Environment & Implemented \\
\bottomrule
\end{tabularx}
\caption{Illustrative A3CG examples showing the aspect--category--action annotation structure.}
\label{tab:a3cg_examples}
\end{table*}

\begin{table*}[!t]
\centering
\captionsetup{font=small}
\caption{\textbf{LLaMA-3-8B — LoRA 8/16}. Abbreviations: CL=Contrastive, TL=Two Losses, G=Gating, L=Lambdas, MGN=MetaGradNorm.}
\label{tab:llama-lora-8-16}
\vspace{0.4em}
\begin{adjustbox}{max width=\textwidth}
\begin{tabular}{lccccccc}
\toprule
\textbf{Type} &
\textbf{(Full)} &
\begin{tabular}{@{}c@{}}\textbf{Seen}\\(1)\end{tabular} &
\begin{tabular}{@{}c@{}}\textbf{Unseen}\\(1)\end{tabular} &
\begin{tabular}{@{}c@{}}\textbf{Seen}\\(2)\end{tabular} &
\begin{tabular}{@{}c@{}}\textbf{Unseen}\\(2)\end{tabular} &
\begin{tabular}{@{}c@{}}\textbf{Seen}\\(3)\end{tabular} &
\begin{tabular}{@{}c@{}}\textbf{Unseen}\\(3)\end{tabular} \\
\midrule
LoRA &
0.5915 & 0.5095 & \textbf{0.4526} & 0.5908 & 0.4612 & 0.6373 & 0.3574 \\
LoRA &
0.5668 & 0.5095 & 0.4526 & 0.6181 & 0.4422 & 0.5972 & 0.3481 \\
LoRA + CL &
0.6195 & 0.5469 & 0.3700 & 0.6217 & 0.4450 & 0.6404 & 0.3529 \\
LoRA + CL &
0.6234 & 0.5588 & 0.3636 & 0.6186 & \textbf{0.4896} & \textbf{0.6965} & 0.3313 \\
LoRA + CL + TL &
0.6044 & 0.5078 & 0.4039 & 0.6000  & 0.3925  & 0.6202  & 0.3676  \\
LoRA + CL + TL &
0.6220 & 0.5594  & 0.3782 & 0.6191 & 0.4258 & 0.6096 & \textbf{0.4000}  \\
LoRA + CL + TL + G &
0.6145 & 0.5530 & 0.4444 & 0.5798 & 0.4400 & 0.6272 & 0.3450 \\
LoRA + CL + TL + G &
\textbf{0.6319} & \textbf{0.5884} & 0.4128 & 0.6290 & 0.4266 & 0.6411 & 0.3556 \\
LoRA + CL + TL + G + L &
0.6241 & 0.5469 & 0.4291 & 0.6048 & 0.4045 & 0.6102 & 0.3554 \\
LoRA + CL + TL + G + L &
0.6241 & 0.5803 & 0.3896 & 0.6309 & 0.4513 & 0.6564 & 0.3770 \\
LoRA + CL + TL + G + L + MGN &
0.6166 & 0.5459 & 0.4054 & \textbf{0.6364} & 0.4523 & 0.5971 & 0.3691 \\
LoRA + CL + TL + G + L + MGN &
0.6075 & 0.5678 & 0.4109 & 0.6346 & 0.4566 & 0.6669 & 0.3665 \\
\bottomrule
\end{tabular}
\end{adjustbox}
\end{table*}

\begin{table*}[!t]
\centering
\captionsetup{font=small}
\caption{\textbf{Gemma-7B — LoRA 8/16}. Abbreviations: CL=Contrastive, TL=Two Losses, G=Gating, L=Lambdas, MGN=MetaGradNorm.}
\label{tab:gemma-lora-8-16}
\vspace{0.4em}
\begin{adjustbox}{max width=\textwidth}
\begin{tabular}{lccccccc}
\toprule
\textbf{Type} &
\textbf{(Full)} &
\begin{tabular}{@{}c@{}}\textbf{Seen}\\(1)\end{tabular} &
\begin{tabular}{@{}c@{}}\textbf{Unseen}\\(1)\end{tabular} &
\begin{tabular}{@{}c@{}}\textbf{Seen}\\(2)\end{tabular} &
\begin{tabular}{@{}c@{}}\textbf{Unseen}\\(2)\end{tabular} &
\begin{tabular}{@{}c@{}}\textbf{Seen}\\(3)\end{tabular} &
\begin{tabular}{@{}c@{}}\textbf{Unseen}\\(3)\end{tabular} \\
\midrule
LoRA &
0.5937 & 0.5747 & 0.4510 & 0.5737 & 0.3985 & 0.6043 & 0.3580 \\
LoRA &
0.5904 & 0.5885 & 0.3874 & 0.6048 & \textbf{0.4845} & 0.5921 & \textbf{0.3921} \\
LoRA + CL &
0.6192 & 0.5794 & 0.2825 & 0.5732 & 0.3888 & 0.5711 & 0.3531 \\
LoRA + CL &
0.6236 & 0.5887 & 0.2953 & 0.5909 & 0.4287 & 0.5979 & 0.3553 \\
LoRA + CL + TL &
0.5978 & \textbf{0.5912} & 0.3627 & 0.5485 & 0.4043 & 0.5934 & 0.3186  \\
LoRA + CL + TL &
\textbf{0.6481} & 0.5741 & 0.3708 & \textbf{0.6382} & 0.4293 & 0.6070 & 0.3543  \\
LoRA + CL + TL + G &
0.6269 & 0.5728 & 0.3519 & 0.5466 & 0.4348 & 0.6068 & 0.3537 \\
LoRA + CL + TL + G &
0.6269 & 0.5798 & \textbf{0.4974} & 0.6226 & 0.4490 & \textbf{0.6403} & 0.3761 \\
LoRA + CL + TL + G + L &
0.6031 & 0.5842 & 0.3486 & 0.5166 & 0.3693 & 0.6073 & 0.3343 \\
LoRA + CL + TL + G + L &
0.6318 & 0.5795 & 0.4016 & 0.6137 & 0.4641 & 0.6159 & 0.3727 \\
LoRA + CL + TL + G + L + MGN &
0.6110 & 0.5714 & 0.3599 & 0.5714 & 0.4213 & 0.6135 & 0.3691 \\
LoRA + CL + TL + G + L + MGN &
0.6379 & 0.5626 & 0.3974 & 0.6363 & 0.4400 & 0.6170 & 0.3497 \\
\bottomrule
\end{tabular}
\end{adjustbox}
\end{table*}

\begin{table*}[!t]
\centering
\captionsetup{font=small}
\caption{\textbf{Mistral-7B — LoRA 8/16}. Abbreviations: CL=Contrastive, TL=Two Losses, G=Gating, L=Lambdas, MGN=MetaGradNorm.}
\label{tab:mistral-lora-8-16}
\vspace{0.4em}
\begin{adjustbox}{max width=\textwidth}
\begin{tabular}{lccccccc}
\toprule
\textbf{Type} &
\textbf{(Full)} &
\begin{tabular}{@{}c@{}}\textbf{Seen}\\(1)\end{tabular} &
\begin{tabular}{@{}c@{}}\textbf{Unseen}\\(1)\end{tabular} &
\begin{tabular}{@{}c@{}}\textbf{Seen}\\(2)\end{tabular} &
\begin{tabular}{@{}c@{}}\textbf{Unseen}\\(2)\end{tabular} &
\begin{tabular}{@{}c@{}}\textbf{Seen}\\(3)\end{tabular} &
\begin{tabular}{@{}c@{}}\textbf{Unseen}\\(3)\end{tabular} \\
\midrule
LoRA &
0.5873 & 0.5394 & \textbf{0.4521} & 0.6248 & 0.5000 & 0.6264 & \textbf{0.3977} \\
LoRA &
0.5963 & \textbf{0.6191} & 0.3880 & 0.6248 & 0.5000 & 0.6264 & 0.3977 \\
LoRA + CL &
0.5587 & 0.5570 & 0.4184 & \textbf{0.6444} & 0.5000 & 0.6425 & 0.3955 \\
LoRA + CL &
0.6102 & 0.6136 & 0.4016 & 0.6380 & 0.4575 & 0.6517 & 0.3866 \\
LoRA + CL + TL &
0.6291 & 0.5828 & 0.3939 & 0.6306 & 0.4463 & 0.6173 & 0.3529 \\
LoRA + CL + TL &
\textbf{0.6397} & 0.5998 & 0.4158 & 0.6247 & 0.4595 & 0.6486  & 0.3278  \\
LoRA + CL + TL + G &
0.6090 & 0.5583 & 0.3830 & 0.6145 & \textbf{0.5087} & \textbf{0.6667} & 0.3930 \\
LoRA + CL + TL + G &
0.6252 & 0.5877 & 0.4191 & 0.6263 & 0.4519 & 0.6572 & 0.3681 \\
LoRA + CL + TL + G + L &
0.6156 & 0.5912 & 0.3956 & 0.6137 & 0.4544 & 0.6145 & 0.3091 \\
LoRA + CL + TL + G + L &
0.6302 & 0.5915 & 0.4303 & 0.6299 & 0.4716 & 0.6590 & 0.3612 \\
LoRA + CL + TL + G + L + MGN &
0.6206 & 0.5794 & 0.3786 & 0.6317 & 0.4361 & 0.6510 & 0.3864 \\
LoRA + CL + TL + G + L + MGN &
0.6347 & 0.5990 & 0.3672 & 0.6130 & 0.4111 & 0.6431 & 0.3618 \\
\bottomrule
\end{tabular}
\end{adjustbox}
\end{table*}

\begin{table*}[!t]
\centering
\captionsetup{font=small}
\caption{\textbf{DeepSeek-7B — LoRA 8/16}. Abbreviations: CL=Contrastive, TL=Two Losses, G=Gating, L=Lambdas, MGN=MetaGradNorm.}
\label{tab:deepseek-lora-8-16}
\vspace{0.4em}
\begin{adjustbox}{max width=\textwidth}
\begin{tabular}{lccccccc}
\toprule
\textbf{Type} &
\textbf{(Full)} &
\begin{tabular}{@{}c@{}}\textbf{Seen}\\(1)\end{tabular} &
\begin{tabular}{@{}c@{}}\textbf{Unseen}\\(1)\end{tabular} &
\begin{tabular}{@{}c@{}}\textbf{Seen}\\(2)\end{tabular} &
\begin{tabular}{@{}c@{}}\textbf{Unseen}\\(2)\end{tabular} &
\begin{tabular}{@{}c@{}}\textbf{Seen}\\(3)\end{tabular} &
\begin{tabular}{@{}c@{}}\textbf{Unseen}\\(3)\end{tabular} \\
\midrule
LoRA &
0.3690 & 0.2727 & 0.2751 & 0.4190 & 0.2902 & 0.3540 & 0.3083 \\
LoRA &
0.3690 & 0.3261 & 0.3082 & 0.4952 & 0.3417 & 0.3902 & 0.3065 \\
LoRA + CL &
0.4434 & 0.3578 & 0.2663 & 0.4698 & 0.3045 & 0.4189 & 0.3219 \\
LoRA + CL &
0.4288 & 0.3862 & 0.2659 & \textbf{0.5284} & \textbf{0.3972} & 0.4501 & 0.3037 \\
LoRA + CL + TL &
0.4051 & 0.3788 & 0.3056 & 0.4474 & 0.3117 & 0.4525 & \textbf{0.3268} \\
LoRA + CL + TL &
0.4371 & 0.3414 & 0.2720 & 0.4802 & 0.3758 & \textbf{0.4800} & 0.3246 \\
LoRA + CL + TL + G &
0.4232 & 0.3572 & 0.2615 & 0.5170 & 0.3549 & 0.3921 & 0.3142 \\
LoRA + CL + TL + G &
0.4366 & 0.3666 & 0.2885 & 0.4825 & 0.3504 & 0.4652 & 0.3130 \\
LoRA + CL + TL + G + L &
0.4408 & 0.3604 & 0.2943 & 0.4747 & 0.3700 & 0.3361 & 0.2853 \\
LoRA + CL + TL + G + L &
\textbf{0.4521} & \textbf{0.4543} & \textbf{0.3144} & 0.4890 & 0.3831 & 0.4475 & 0.2887 \\
LoRA + CL + TL + G + L + MGN &
0.4354 & 0.3127 & 0.2932 & 0.5021 & 0.3803 & 0.3721 & 0.2891 \\
LoRA + CL + TL + G + L + MGN &
0.4405 & 0.4425 & 0.2953 & 0.5204 & 0.3869 & 0.4547 & 0.3053 \\
\bottomrule
\end{tabular}
\end{adjustbox}
\end{table*}

\begin{table*}[!t]
\centering
\captionsetup{font=small}
\caption{\textbf{Qwen2.5-7B — LoRA 8/16}. Abbreviations: CL=Contrastive, TL=Two Losses, G=Gating, L=Lambdas, MGN=MetaGradNorm.}
\label{tab:qwen-lora-8-16}
\vspace{0.4em}
\begin{adjustbox}{max width=\textwidth}
\begin{tabular}{lccccccc}
\toprule
\textbf{Type} &
\textbf{(Full)} &
\begin{tabular}{@{}c@{}}\textbf{Seen}\\(1)\end{tabular} &
\begin{tabular}{@{}c@{}}\textbf{Unseen}\\(1)\end{tabular} &
\begin{tabular}{@{}c@{}}\textbf{Seen}\\(2)\end{tabular} &
\begin{tabular}{@{}c@{}}\textbf{Unseen}\\(2)\end{tabular} &
\begin{tabular}{@{}c@{}}\textbf{Seen}\\(3)\end{tabular} &
\begin{tabular}{@{}c@{}}\textbf{Unseen}\\(3)\end{tabular} \\
\midrule
LoRA &
0.5874 & 0.5203 & 0.4079 & 0.5931 & 0.4255 & 0.5405 & 0.3939 \\
LoRA &
0.5906 & 0.5203 & 0.4079 & 0.6161 & 0.4636 & 0.6109 & 0.4022 \\
LoRA + CL &
\textbf{0.6262} & 0.5116 & 0.3413 & 0.5774 & 0.4417 & 0.5929 & 0.2963 \\
LoRA + CL &
0.6099 & 0.5309 & 0.3995 & 0.5788 & 0.3864 & 0.6170 & 0.3703 \\
LoRA + CL + TL &
0.6132 & 0.5330 & 0.3802 & 0.5954 & 0.4238 & 0.6254 & 0.3516 \\
LoRA + CL + TL &
0.6047 & 0.5227 & 0.3251 & 0.6301 & \textbf{0.4715} & \textbf{0.6471} & 0.3865 \\
LoRA + CL + TL + G &
0.5734 & 0.5483 & 0.4037 & 0.5910 & 0.4698 & 0.6121 & 0.3771 \\
LoRA + CL + TL + G &
0.6193 & 0.5593 & 0.3449 & 0.6144 & 0.4136 & 0.6182 & \textbf{0.4218} \\
LoRA + CL + TL + G + L &
0.6155 & 0.5580 & \textbf{0.4216} & \textbf{0.6340} & 0.4272 & 0.6117 & 0.3877 \\
LoRA + CL + TL + G + L &
0.5934  & \textbf{0.5669} & 0.3767 & 0.5990 & 0.4054 & 0.6371 & 0.3949 \\
LoRA + CL + TL + G + L + MGN &
0.6160 & 0.5236 & 0.3347 & 0.5980 & 0.4291 & 0.6198 & 0.3711 \\
LoRA + CL + TL + G + L + MGN &
0.6006 & 0.5465 & 0.3440 & 0.6185 & 0.4387 & 0.6173 & 0.3902 \\
\bottomrule
\end{tabular}
\end{adjustbox}
\end{table*}

\begin{table*}[ht]
\centering
\caption{LLM Results: Fine-tuned models (LoRA) and zero-shot/few-shot baselines}
\adjustbox{max width=\textwidth}{
\begin{tabular}{|l|c|cc|cc|cc|c|c|c|}
\hline
\multirow{2}{*}{\textbf{Method}} & \textbf{Full} & \multicolumn{2}{c|}{\textbf{Fold 1}} & \multicolumn{2}{c|}{\textbf{Fold 2}} & \multicolumn{2}{c|}{\textbf{Fold 3}} & \multirow{2}{*}{\textbf{S Avg}} & \multirow{2}{*}{\textbf{US Avg}} & \multirow{2}{*}{$\Delta$} \\
\cline{3-8}
 & \textbf{Dataset} & S & US & S & US & S & US & & & \\
\hline
\multicolumn{11}{|c|}{\textit{Fine-tuned (LoRA)}} \\
\hline
LLaMA-3-8B & 0.5668 & 0.5095 & \underline{0.4626} & 0.6181 & 0.4422 & 0.5972 & 0.3481 & 0.5749 & 0.4176 & -0.1573 \\
LLaMA-3-8B + COGLM & 0.6075 & 0.5678 & 0.4109 & 0.6346 & 0.4566 & \textbf{\underline{0.6669}} & 0.3665 & 0.6231 & 0.4113 & -0.2118 \\
\hline
Mistral-7B & 0.5963 & \textbf{\underline{0.6191}} & 0.3880 & 0.6248 & \textbf{\underline{0.5000}} & 0.6264 & 0.3977 & \textbf{\underline{0.6234}} & \textbf{\underline{0.4286}} & -0.1948 \\
Mistral-7B + COGLM & 0.6206 & 0.5794 & 0.3786 & 0.6317 & 0.4361 & 0.6510 & 0.3864 & 0.6207 & 0.4004 & -0.2203 \\
\hline
Gemma-7B & 0.5937 & 0.5747 & 0.4510 & 0.5737 & 0.3985 & 0.6043 & 0.3580 & 0.5842 & 0.4025 & -0.1817 \\
Gemma-7B + COGLM & \textbf{\underline{0.6379}} & 0.5626 & 0.3974 & \textbf{\underline{0.6363}} & 0.4400 & 0.6170 & 0.3497 & 0.6053 & 0.3957 & -0.2096 \\
\hline
DeepSeek-7B & 0.3690 & 0.3261 & 0.3082 & 0.4952 & 0.3417 & 0.3902 & 0.3065 & 0.4038 & 0.3188 & -0.0850 \\
DeepSeek-7B + COGLM & 0.4405 & 0.4425 & 0.2953 & 0.5204 & 0.3869 & 0.4547 & 0.3053 & 0.4725 & 0.3292 & -0.1433 \\
\hline
Qwen2.5-7B & 0.5906 & 0.5203 & 0.4079 & 0.6161 & 0.4636 & 0.6109 & \textbf{\underline{0.4022}} & 0.5824 & 0.4246 & -0.1578 \\
Qwen2.5-7B + COGLM & 0.6006 & 0.5465 & 0.3440 & 0.6185 & 0.4387 & 0.6173 & 0.3902 & 0.5941 & 0.3910 & -0.2031 \\
\hline
\multicolumn{11}{|c|}{\textit{Zero-shot / Few-shot}} \\
\hline
GPT-4o & 0.2979 & 0.3161 & 0.4298 & 0.4000 & 0.3251 & 0.3558 & 0.3235 & 0.3573 & 0.3595 & +0.0022 \\
GPT-4o + FS* & 0.3569 & 0.3908 & 0.4668 & 0.3912 & 0.4110 & 0.4005 & 0.3346 & 0.3942 & 0.4041 & +0.0099 \\
\hline
Claude 3.5 Sonnet & 0.3770 & 0.3671 & 0.3944 & 0.4111 & 0.3688 & 0.4159 & 0.3822 & 0.3980 & 0.3818 & -0.0162 \\
Claude 3.5 Sonnet + FS* & \underline{0.4211} & \underline{0.4062} & 0.4627 & 0.4318 & 0.4035 & 0.4504 & \underline{0.3948} & 0.4295 & \underline{0.4203} & -0.0092 \\
\hline
Llama 3 (70B) & 0.2015 & 0.1797 & 0.2524 & 0.2366 & 0.1833 & 0.2231 & 0.1843 & 0.2131 & 0.2067 & -0.0064 \\
Llama 3 (70B) + FS* & 0.2982 & 0.2503 & 0.3811 & 0.3349 & 0.3330 & 0.3315 & 0.2665 & 0.3056 & 0.3269 & +0.0213 \\
\hline
DeepSeek V3 & 0.4016 & 0.3806 & \textbf{\underline{0.4818}} & \underline{0.4527} & 0.4023 & 0.4246 & 0.3484 & 0.4196 & 0.4108 & -0.0088 \\
DeepSeek V3 + FS* & 0.3563 & 0.3885 & 0.3016 & 0.4418 & \underline{0.4451} & \underline{0.4662} & 0.3443 & \underline{0.4322} & 0.3637 & -0.0685 \\
\hline
\end{tabular}
}
\label{tab:llm_results_bis}
\end{table*}

\begin{table*}[!t]
\centering
\captionsetup{font=small}
\caption{\textbf{T5}. Abbreviations: FT=Fine Tune, CL=Contrastive, TL=Two Losses, G=Gating, MGN=MetaGradNorm.}
\label{tab:t5}
\vspace{0.4em}
\begin{adjustbox}{max width=\textwidth}
\begin{tabular}{lccccccc}
\toprule
\textbf{Type} &
\textbf{(Full)} &
\begin{tabular}{@{}c@{}}\textbf{Seen}\\(1)\end{tabular} &
\begin{tabular}{@{}c@{}}\textbf{Unseen}\\(1)\end{tabular} &
\begin{tabular}{@{}c@{}}\textbf{Seen}\\(2)\end{tabular} &
\begin{tabular}{@{}c@{}}\textbf{Unseen}\\(2)\end{tabular} &
\begin{tabular}{@{}c@{}}\textbf{Seen}\\(3)\end{tabular} &
\begin{tabular}{@{}c@{}}\textbf{Unseen}\\(3)\end{tabular} \\
\midrule
FT &
0.7160 & 0.5676 & 0.4863 & 0.6683 & 0.4947 & \textbf{0.6994} & 0.3589 \\
FT + CL &
0.6955 & 0.5812 & 0.4657 & 0.6505 & 0.4921 & 0.6919 & 0.3660 \\
FT + CL + TL &
0.6943 & 0.5925 & 0.4332 & 0.6746 & \textbf{0.5141} & 0.6910 & 0.4353 \\
FT + CL + TL + G + L &
0.7172 & 0.6220 & 0.4967 & 0.6516 & 0.4988 & 0.6818 & 0.4508 \\
FT + CL + TL + G + MGN &
\textbf{0.7240} & \textbf{0.6364} & \textbf{0.5211} & \textbf{0.6997} & 0.4857 & 0.6915 & \textbf{0.4644} \\
\bottomrule
\end{tabular}
\end{adjustbox}
\end{table*}

\begin{table*}[ht]
\centering
\caption{T5 Results: Comparison between our method and A3CG baseline}
\adjustbox{max width=\textwidth}{
\begin{tabular}{|l|c|cc|cc|cc|c|c|c|}
\hline
\multirow{2}{*}{\textbf{Method}} & \textbf{Full} & \multicolumn{2}{c|}{\textbf{Fold 1}} & \multicolumn{2}{c|}{\textbf{Fold 2}} & \multicolumn{2}{c|}{\textbf{Fold 3}} & \multirow{2}{*}{\textbf{S Avg}} & \multirow{2}{*}{\textbf{US Avg}} & \multirow{2}{*}{$\Delta$} \\
\cline{3-8}
 & \textbf{Dataset} & S & US & S & US & S & US & & & \\
\hline
\multicolumn{11}{|c|}{\textit{Ours}} \\
\hline
T5 & 71.60 & 56.76 & 48.63 & 66.83 & \underline{49.47} & \underline{69.94} & 35.89 & 64.51 & 44.66 & -19.85 \\
T5 + COGLM & \textbf{\underline{72.40}} & \textbf{\underline{63.64}} & \textbf{\underline{52.11}} & \textbf{\underline{69.97}} & 48.57 & 69.15 & \underline{46.44} & \textbf{\underline{67.59}} & \textbf{\underline{49.04}} & -18.55 \\
\hline
\multicolumn{11}{|c|}{\textit{A3CG Baselines}} \\
\hline
T5 & 70.48 & 57.85 & 43.03 & 68.90 & 45.74 & 67.94 & 34.59 & 64.90 & 41.12 & -23.78 \\
BERT-ST & 43.19 & 39.56 & 25.25 & 39.00 & 22.92 & 43.97 & 30.01 & 40.84 & 26.06 & -14.78 \\
T5 + CL & \underline{71.12} & \underline{62.96} & \underline{46.97} & \underline{69.76} & 46.67 & \underline{67.99} & 38.33 & \underline{66.90} & \underline{43.99} & -22.91 \\
T5 + AL & 69.27 & 61.24 & 39.62 & 66.91 & \underline{47.02} & 65.17 & \underline{41.82} & 64.44 & 42.82 & -21.62 \\
BERT-ST + CL & 68.53 & 60.00 & 37.06 & 69.22 & 41.78 & 58.04 & 34.94 & 62.42 & 37.93 & -24.49 \\
BERT-ST + AL & 24.30 & 37.75 & 27.05 & 27.13 & 25.57 & 34.70 & 26.11 & 33.19 & 26.24 & -6.95 \\
\hline
\multicolumn{11}{|c|}{\textit{Other ABSA Methods}} \\
\hline
InstructABSA & 69.47 & 60.23 & 37.53 & 64.73 & 49.76 & 64.14 & 47.38 & 63.03 & 44.89 & -18.14 \\
IT-RER-ABSA & 69.20 & 57.70 & 41.81 & 66.02 & 48.87 & 68.83 & 39.10 & 64.18 & 43.26 & -20.96 \\
GRACE & 67.09 & 60.87 & \underline{50.08} & 63.10 & 44.33 & 61.92 & \textbf{\underline{48.12}} & 61.96 & \underline{47.51} & -14.45 \\
CONTRASTE & \underline{71.33} & \underline{61.26} & 48.14 & \underline{69.81} & \textbf{\underline{50.30}} & \textbf{\underline{71.34}} & 40.58 & \underline{67.47} & 46.34 & -21.13 \\
\hline
\end{tabular}
}
\label{tab:t5_results_bis}
\end{table*}

\begin{table*}[!t]
\centering
\captionsetup{font=small}
\caption{\textbf{LLaMA-3-8B (Hyperparameter tuning)}. Abbreviations: CL=Contrastive, TL=Two Losses, G=Gating, L=Lambdas, MGN=MetaGradNorm.}
\label{tab:llama-ablation}
\vspace{0.4em}
\begin{tabular}{lcc}
\toprule
\textbf{Type} & \textbf{Seen (1)} & \textbf{Unseen (1)} \\
\midrule
LoRA & 0.5380 & 0.4439 \\
LoRA & 0.5095 & 0.4526 \\
LoRA & 0.5104 & 0.4830 \\
LoRA & 0.5193 & \textbf{0.4884} \\
LoRA + CL & 0.4513 & 0.3551 \\
LoRA + CL & 0.5469 & 0.3700 \\
LoRA + CL & 0.5365 & 0.3825 \\
LoRA + CL + TL & 0.5078 & 0.4039 \\
LoRA + CL + TL & 0.5362 & 0.3370 \\
LoRA + CL + TL & \textbf{0.5794} & 0.3812 \\
LoRA + CL + TL + G & 0.5489 & 0.4423 \\
LoRA + CL + TL + G & 0.5530 & 0.4444 \\
LoRA + CL + TL + G & 0.5188 & 0.4510 \\
LoRA + CL + TL + G + L & 0.5308 & 0.4525 \\
LoRA + CL + TL + G + L & 0.5469 & 0.4291 \\
LoRA + CL + TL + G + L & 0.5622 & 0.4239 \\
LoRA + CL + TL + G + L + MGN & 0.5459 & 0.4054 \\
LoRA + CL + TL + G + L + MGN & 0.5501 & 0.4190 \\
LoRA + CL + TL + G + L + MGN & 0.5223 & 0.4128 \\
\bottomrule
\end{tabular}
\end{table*}

\end{document}